\documentclass[runningheads]{llncs}

\usepackage{eccv}

\usepackage{eccvabbrv}

\usepackage{graphicx}
\usepackage{booktabs}

\usepackage{amsmath,amssymb}
\usepackage{arydshln}
\usepackage{multirow}
\usepackage{tabularx}
\usepackage{makecell}
\usepackage{wrapfig}
\usepackage{colortbl}
\usepackage{bm}
\newcolumntype{C}{>{\centering\arraybackslash}X}

\usepackage[accsupp]{axessibility}  % Improves PDF readability for those with disabilities.

\usepackage{hyperref}

\usepackage{orcidlink}

\begin{document}

% ---------------------------------------------------------------
\title{Pixel Ignores, Superpixel Sees: Adverse Weather Image Restoration via Semantic-Center SSM} 

\titlerunning{Adverse Weather Image Restoration via Semantic-Center SSM}

\author{Dayu Li\inst{3}\orcidlink{0009-0005-8953-642X} \and
Shihao Zhou\inst{1,3}\orcidlink{0000-0002-9202-9761} \and
Leizhi Shu\inst{4}\orcidlink{0009-0004-9416-7078} \and
Jin Wu\inst{4}\orcidlink{0000-0001-5930-4170} \and
Chi Man Vong\inst{5}\orcidlink{0000-0001-7997-8279} \and
Jufeng Yang\inst{1,2,3}\orcidlink{0000-0003-0219-3443}\thanks{Corresponding author.}}

% 缩写名称 (Running head)
\authorrunning{D.~Li et al.}

% 机构信息
\institute{
Nankai International Advanced Research Institute (SHENZHEN$\cdot$FUTIAN) \\
\and
Peng Cheng Laboratory \\
\and
College of Computer Science, Nankai University \\
\and
University of Science and Technology Beijing \\
\and
Department of Computer and Information Science, University of Macau
}

\maketitle

\begin{abstract}
  Adverse weather image restoration aims to recover clear visibility from degraded images in complex weather conditions. 
  Existing works attempt to address this problem by modeling relationships between pixels, however, this paradigm defies the spatially non-uniformity fact of degradations and learns non-discriminative features from semantic-conflict regions. 
  In this paper, we propose SSR, a \textbf{S}emantic-center guilded \textbf{S}tate space model for image \textbf{R}estoration.
  The key idea of SSR is to shift the conventional scanning strategy of pixel-serial to semantic-guilded one.
  Specifically, we introduce a Superpixel-guided Selective Scan Mechanism ($\text{S}^3$M), which first partitions the image into perceptually coherent regions via superpixel clustering and then performs relations modeling within the semantic-related regions.  
  Moreover, a Region-level Gating Mechanism (RGM) is developed to perform intra-region calibration by modulating degradation outliers within each semantic superpixel unit along the channel dimension. 
  Extensive experiments on \textbf{6} well-established benchmarks demonstrate that SSR performs favorably against state-of-the-art models with competitive computational cost.
  The source code is publicly available at \url{https://github.com/LIDAYU-DayuLi/SSR}.
\end{abstract}

\section{Introduction}

Adverse weather conditions, such as rain, haze, and snow, pose challenges to vision systems deployed in real-world environments. Serious degradations distort scene structures and corrupt visual cues in a spatially non-uniform way, leading to performance degradation in downstream applications~\cite{zhao2024revisiting,gasperini2023robust,sakaridis2021acdc}.
Intuitively, developing a unified image restoration model for handling different adverse weather conditions provides a plausible solution.

Previous restoration methods~\cite{he2010single,rudin1992nonlinear} primarily rely on hand-crafted priors to approximate the physical process of degradation.
Unfortunately, these simplified assumptions often fall short in capturing the diverse and non-linear patterns observed in real-world weather.
To address this, deep learning-based approaches~\cite{fu2017removing,liu2018desnownet} have shifted the paradigm toward data-driven one, achieving superior performance by learning complex mappings from large-scale paired datasets. 
Nevertheless, most existing models are constrained to task-specific settings (e.g., only deraining or dehazing), which leads to limited generalization capability when encountering different or hybrid degradations under in-the-wild environments.

\begin{figure}[t!]
    \centering
    \begin{minipage}{0.24\linewidth}
        \centering
        \includegraphics[width=\linewidth]{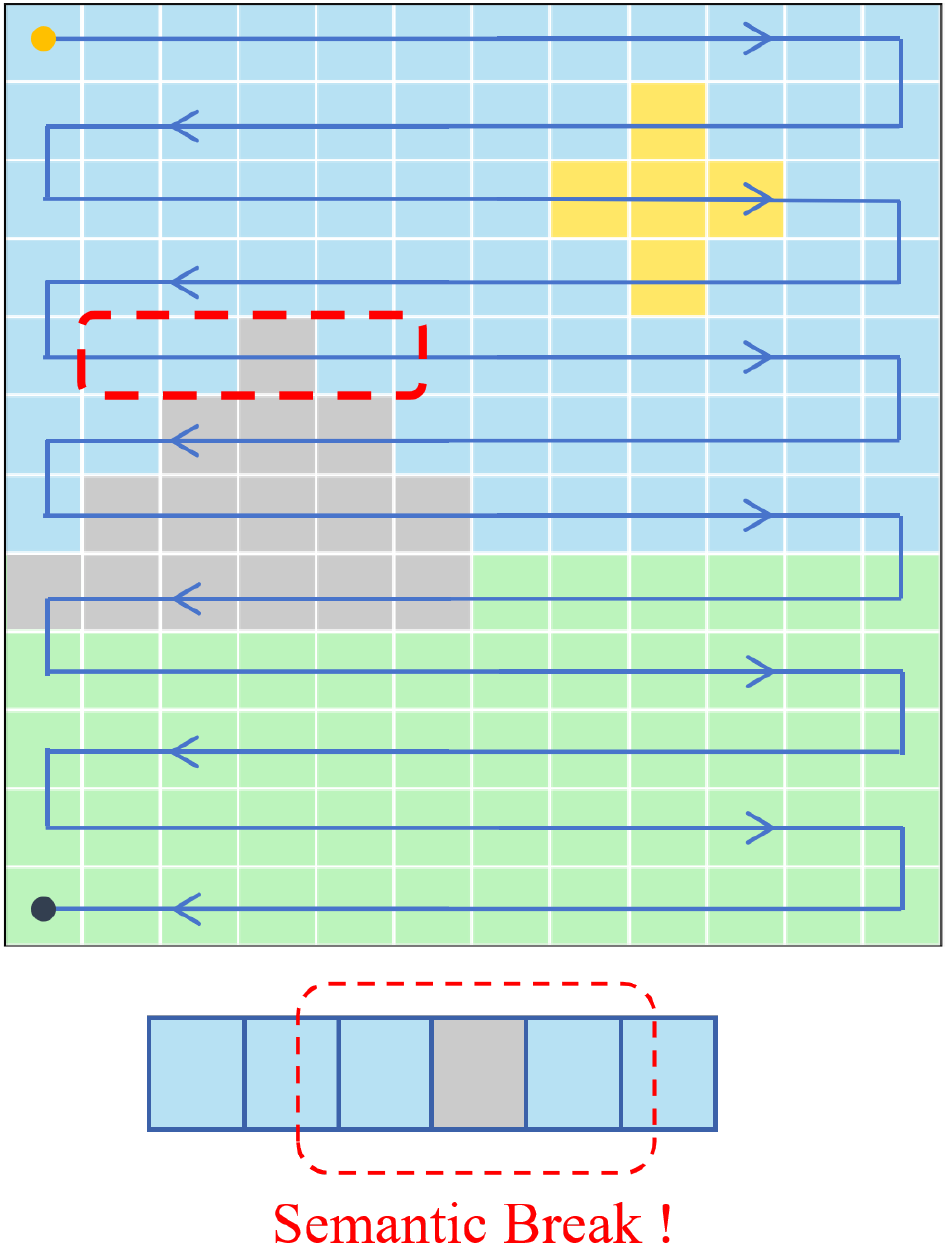}
        \\ \vspace{1mm}
        {\scriptsize (a) Raster\cite{liu2024vmamba}}
    \end{minipage}\hfill
    \begin{minipage}{0.24\linewidth}
        \centering
        \includegraphics[width=\linewidth]{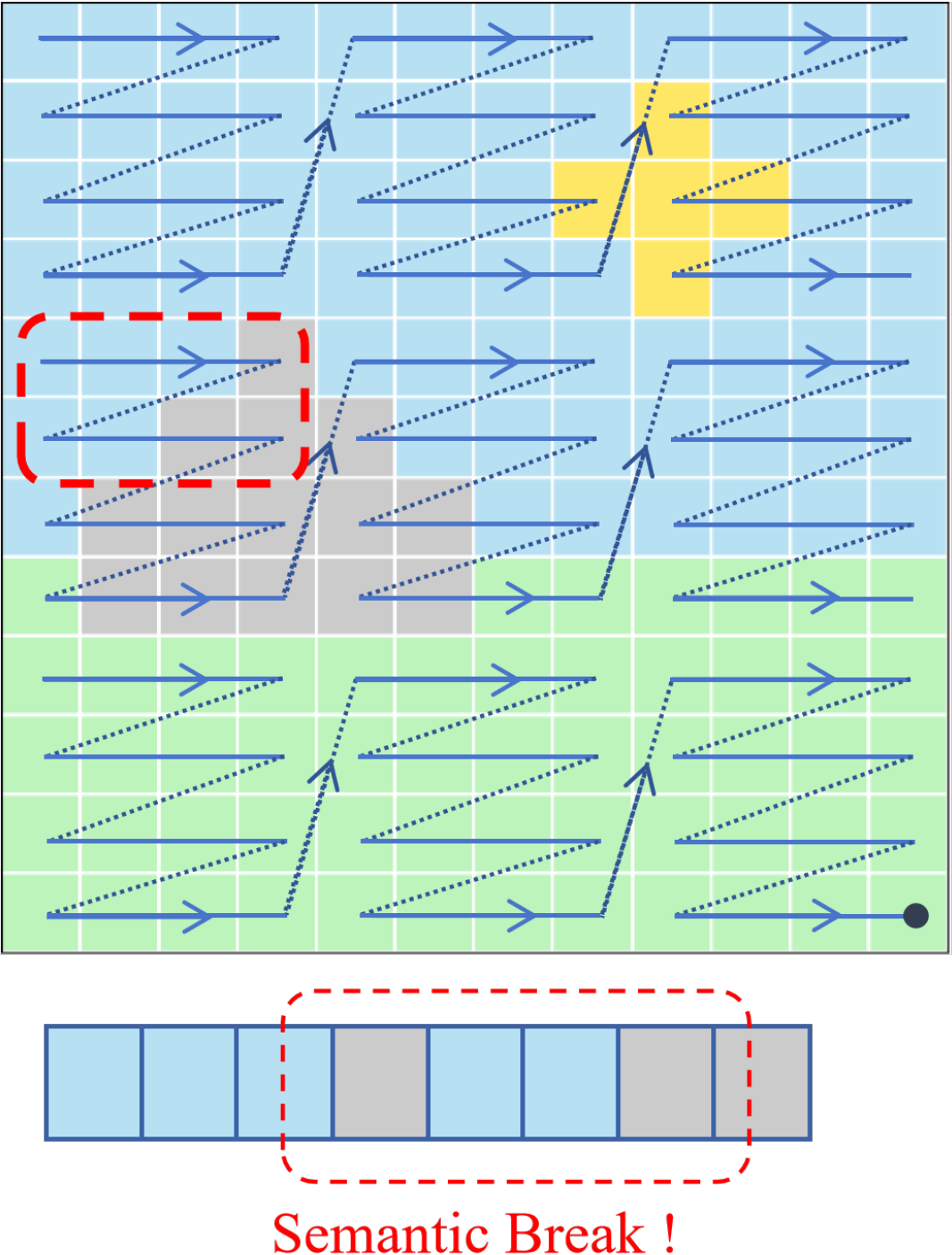}
        \\ \vspace{1mm}
        {\scriptsize (b) Local Window\cite{huang2024localmamba}}
    \end{minipage}\hfill
    \begin{minipage}{0.24\linewidth}
        \centering
        \includegraphics[width=\linewidth]{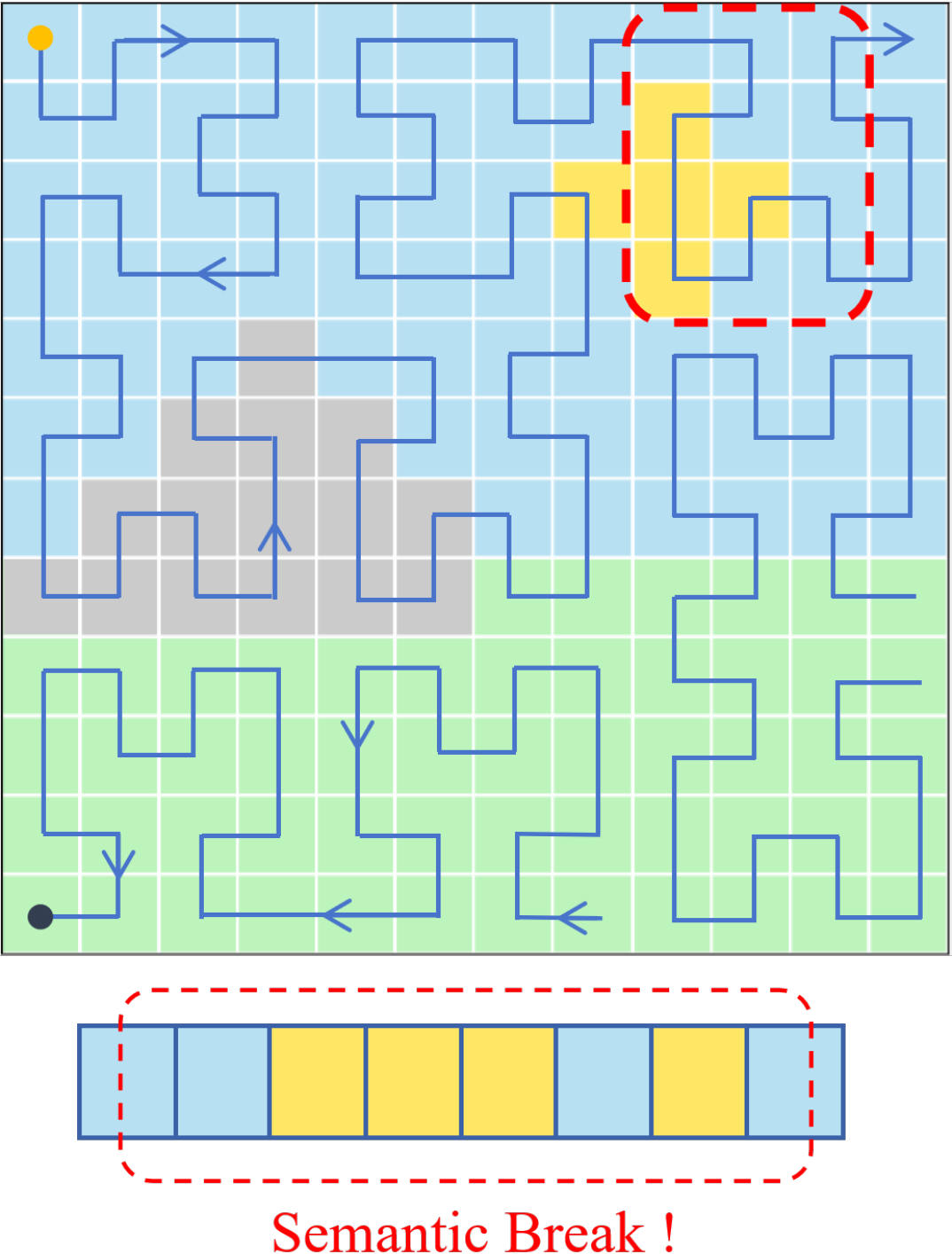}
        \\ \vspace{1mm}
        {\scriptsize (c) Hilbert\cite{wu2024rainmamba}}
    \end{minipage}\hfill
    \begin{minipage}{0.24\linewidth}
        \centering
        \includegraphics[width=\linewidth]{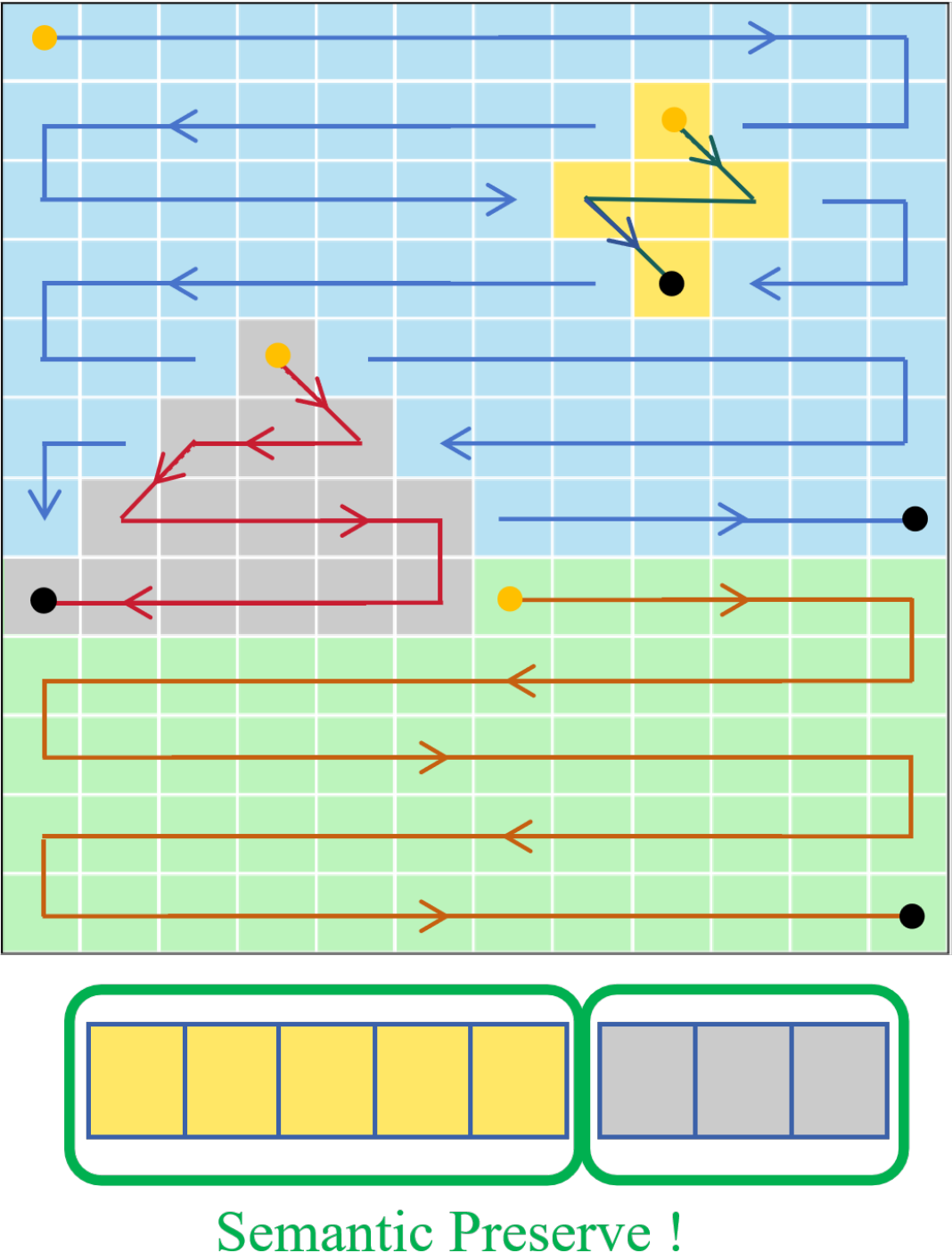}
        \\ \vspace{1mm}
        {\scriptsize (d) $\text{S}^3$M(ours)}
    \end{minipage}
    \caption{
        \textbf{Comparison of scanning mechanisms.} 
        (a-c) Existing predefined scan strategies fall into a content-agnostic paradigm, and may learn non-discriminative features from semantic-conflict regions.
        (d) Our Superpixel-guided Selective Scan Mechanism (S$^{3}$M) Scan constrains the trajectory within perceptually coherent regions, effectively preventing information leakage and ensuring structure-aware state evolution.
    }
    \label{fig:intro}
\end{figure}

To address the diverse and complex degradations in adverse weather, All-in-one Image Restoration (AIR) frameworks aim to restore images using a single unified model. 
Early CNN-based methods~\cite{li2020all,li2022all} are restricted within limited receptive fields, 
while Transformer-based architectures~\cite{zhu2024mwformer,valanarasu2022transweather} achieve global modeling but incur quadratic computational overhead.
More recently, State Space Models (SSMs)~\cite{gu2024mamba,liu2024vmamba}  have emerged as a promising alternative, offering linear-complexity long-range dependency modeling. 
However, directly applying SSMs to the AIR task presents a critical challenge.
For the vision tasks, the SSMs rely on flattening 2D images into 1D sequences for recursive state propagation. 
The vanilla SSMs adopt predefined and geometry-based scanning trajectories, such as standard raster scans\cite{liu2024vmamba} (Fig.~\ref{fig:intro}a), window partitioning~\cite{huang2024localmamba} (Fig.~\ref{fig:intro}b), or hilbert curves~\cite{wu2024rainmamba} (Fig.~\ref{fig:intro}c). 
These attempts achieve advanced performance boosts with limited computational cost, but they are hindered from generating better results by ignoring image semantic content.
SSMs rely on sequential state accumulation to model representation, while these predefined trajectories inevitably include semantically conflicting regions as continuous sequences (e.g., patches cross object boundaries).
These fields are not ideally suited for the SSMs to learn informative features, and heavy mixed degradations further impede the models from achieving satisfactory restoration.

In light of this, we propose the \textbf{S}emantic-center guided \textbf{S}tate space model for \textbf{R}estoration (SSR) to advance the content-agnostic scanning manner in current methods. 
Instead of relying on rigid geometric paths, SSR utilizes superpixels (Fig.~\ref{fig:intro}d) as perceptually coherent units to dynamically construct structure-aware scanning trajectories. This adaptive routing strictly adheres to semantic boundaries, effectively preventing the leakage of degradation noise into clean regions. Furthermore, we introduce a Region-level Gating Mechanism (RGM) to perform intra-region calibration. By adaptively modulating degradation outliers within each semantic superpixel unit along the channel dimension, our framework provides a more interpretable, efficient, and content-adaptive SSM-based model tailored for universal weather restoration.

Our main contributions are summarized as follows.
\begin{itemize}
    \item We propose SSR, a \textbf{S}emantic-center guided \textbf{S}tate space model for image \textbf{R}estoration, which shifts the conventional scanning paradigm of SSMs from pixel serial to superpixel-based semantic-guided one, effectively preventing learning noisy features from the semantic conflict regions.
    \item We introduce the Superpixel-guided Selective Scan Mechanism ($\text{S}^3$M) to model relations within perceptually coherent regions, alongside a Region-level Gating Mechanism (RGM) that performs intra-region calibration by modulating degradation outliers along the channel dimension.
    \item Extensive experiments on \textbf{6} well-established benchmarks demonstrate that SSR performs favorably against state-of-the-art models with competitive computational cost.
\end{itemize}

\section{Related Work}

\textbf{Task-specific image restoration.}
Early work in this field primarily relied on image decomposition or statistical priors (e.g., sparse representation for rain streak removal~\cite{jiang2017novel,kang2011automatic} and the Dark Channel Prior for haze removal~\cite{he2010single}). 
With the widespread adoption of deep learning, the focus shifted towards data-driven modeling, leading to the development of various specialized network architectures. 
such as polarization-aided transformers~\cite{chen2025polar} and detail separation networks~\cite{fu2017removing}, recurrent aggregation networks~\cite{li2018recurrent}, and uncertainty-guided multi-scale designs~\cite{yasarla2019uncertainty} for rain streak removal; 
attention-guided generative adversarial networks~\cite{qian2018attentive} and dual attention models~\cite{zhang2021dual} for raindrop removal; 
transmission map estimation models~\cite{yang2024depth} along with contrastive learning frameworks~\cite{wu2021contrastive} for dehazing; 
and periodicity-aware frameworks for burst flicker removal~\cite{qu2026flickerformer}.
On the other hand,
% Although strong dedicated priors are less common for degradations like snow, 
researchers have also developed effective desnowing models by incorporating context-aware designs~\cite{liu2018desnownet} or leveraging semantic and depth priors~\cite{zhang2021deep}.
More recent solutions are often built upon general CNN backbones~\cite{cui2023image,cui2024revitalizing} or Transformers~\cite{zhou2025learning,zhou2025devil,zhou2026rethinking}. 
However, this specialization in a single degradation type impedes generalization across diverse weather conditions, thereby complicating practical deployment with multiple dedicated models and increased overhead.

\textbf{All-in-one weather removal.}
In recent years, unified all-weather image restoration models have emerged as a significant research direction, aiming to handle multiple weather degradations with a single model. 
Early efforts primarily focused on exploring generic network architectures~\cite{li2020all} or leveraging the global modeling capability of Transformers to construct foundational models~\cite{valanarasu2022transweather}. 
As research has progressed, more diverse approaches have been introduced to enhance model generalization and adaptability, 
including disentangling weather-general and weather-specific features~\cite{zhu2023learning}, adopting generative frameworks such as diffusion models~\cite{ozdenizci2023restoring}, 
integrating learned codebook priors~\cite{ye2023adverse}, and incorporating global image statistics as priors within Transformer-based models like Histoformer~\cite{peng2024histoformer}.
Moreover, some works have explored strategies such as instruction-based natural language guidance~\cite{conde2024instructir} and prompt-encoded degradation conditions~\cite{potlapalli2023promptir}. 
At the same time, efficient long-sequence modeling architectures, represented by State Space Models (SSMs), have been introduced to this task, offering a new approach for modeling long-range degradation dependencies through their linear complexity and selective mechanisms~\cite{wang2025modem}. 
Nevertheless, accurately modeling diverse and spatially heterogeneous weather degradations remains a core challenge in the field.

\textbf{Mamba-based image restoration.}
Recent advances in SSMs have introduced an efficient approach for modeling long-range dependencies in image restoration, leveraging linear complexity and selective scanning mechanisms~\cite{gu2024mamba}. 
General frameworks such as MambaIR~\cite{guo2024mambair} and VMambaIR~\cite{shi2025vmambair} have validated the potential of this architecture as a foundational model for image restoration. 
Concurrently, research on task-specific designs has proliferated, spanning areas including deraining~\cite{zou2024freqmamba,yamashita2024image,li2025fouriermamba}, dehazing~\cite{zheng2024u,zhou2024rsdehamba}, deblurring~\cite{kong2025efficient}, and low-light enhancement~\cite{bai2024retinexmamba,weng2024mamballie}. 
For instance, RainMamba~\cite{wu2024rainmamba} incorporates a Hilbert-order scanning mechanism to enhance the preservation of local structures, 
while Wave-Mamba~\cite{zou2024wave} integrates wavelet transforms to optimize feature representation in the frequency domain. 
Furthermore, studies such as Vision Mamba~\cite{zhu2024vision} and Mamba-ND~\cite{li2024mamba} have extended SSMs into pure visual architectures and multi-dimensional data processing frameworks,improving model capabilities in dense prediction and multi-dimensional signal modeling. 
These developments collectively underscore the ongoing evolution of scanning strategies within SSM frameworks, highlighting their critical role in capturing spatially varying degradations and establishing a foundation for further advances in sequential modeling for image restoration.

\section{Methodology}

\subsection{Overall Architecture}
\label{sec:overall_arch}

\begin{figure*}[t!]
    \centering
    \includegraphics[width=\linewidth]{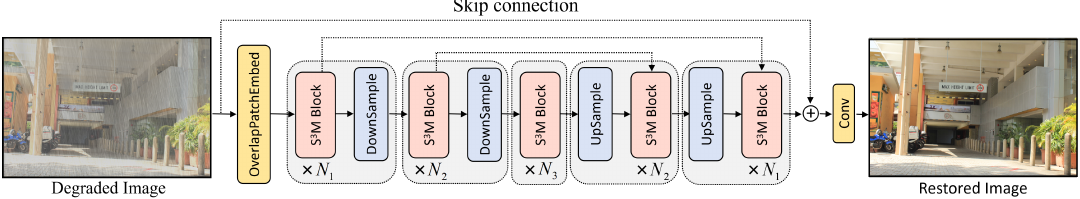}
    \vspace{-0.2cm}
    \caption{
        Overall architecture of the proposed Semantic-center Guided State Space Model \textbf{(SSR)}. 
        Our model adopts a hierarchical U-shaped encoder-decoder structure to capture multi-scale features for unified weather restoration. 
        The backbone consists of multiple $\text{S}^3$M Blocks, which integrate the {Superpixel-guided Selective Scan Mechanism ($\text{S}^3$M)} and a {Region-level Gating Mechanism} (\textbf{RGM}) to perform intra-region calibration by modulating degradation outliers along the channel dimension.
        Skip connections are employed to fuse multi-scale encoder features with upsampled decoder features.
    }
    \label{fig:arch} 
\end{figure*}

The overall architecture of SSR follows a hierarchical encoder-decoder framework (Fig.~\ref{fig:arch}). Given a degraded image $I_{in} \in \mathbb{R}^{H \times W \times 3}$, a $3 \times 3$ convolutional layer is first employed to extract shallow features $F_s \in \mathbb{R}^{H \times W \times C}$, where $C$ denotes the number of feature channels. These features are then fed into a 3-level symmetric encoder-decoder network. Each level consists of several $\text{S}^3$M Blocks (see \cref{sec:s3m_block}). For the level-$l$ encoder/decoder ($l \in \{1, 2, 3\}$), the intermediate features $F_{enc}^l / F_{dec}^l \in \mathbb{R}^{\frac{H}{2^{l-1}} \times \frac{W}{2^{l-1}} \times 2^{l-1}C}$ are progressively processed. We utilize bilinear interpolation and $1 \times 1$ convolutions for downsampling and upsampling, complemented by skip connections between corresponding levels to fuse multi-scale features. Finally, a $3 \times 3$ convolution is applied to generate the restoration residual $R \in \mathbb{R}^{H \times W \times 3}$. The final restored image $I_{rec}$ is obtained by:
\begin{equation}
    I_{rec} = I_{in} + R = I_{in} + \mathcal{F}_{SSR}(I_{in}),
\end{equation}
where $\mathcal{F}_{SSR}(\cdot)$ denotes the proposed SSR network.

\subsection{$\text{S}^3$M Block}
\label{sec:s3m_block}

\begin{figure*}[t!]
    \centering
    \includegraphics[width=\linewidth]{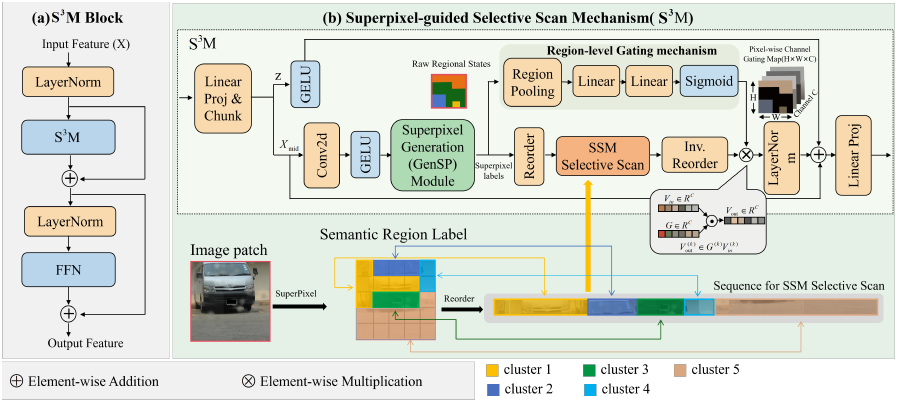}
    \vspace{-0.2cm}
    \caption{
        Structure of the $\text{S}^3$M Block. 
        (a) The $\text{S}^3$M block enables structure-aware feature transitions by integrating the Superpixel-guided Selective Scan Mechanism ($\text{S}^3$M) and a Feed-Forward Network (FFN). 
        (b) The $\text{S}^3$M incorporates two synergistic innovations: Superpixel-guided Scanning, which constrains state propagation within perceptually coherent regions to prevent learning non-discriminative features from semantic-conflict areas, and Region-level Gating, which performs intra-region calibration by modulating degradation outliers within each semantic unit along the channel dimension.
    }
    \label{fig:s3m_block} 
\end{figure*}

The core building block of our encoder and decoder is the $\text{S}^3$M Block (Fig.~\ref{fig:s3m_block}). Given an input feature $X \in \mathbb{R}^{H \times W \times C}$, the block consists of a spatial token mixer and a channel-wise feed-forward network:
\begin{equation}
    X' = \text{S}^3\text{M}(\text{LN}(X)) + X,
\end{equation}
\begin{equation}
    X_{out} = \text{FFN}(\text{LN}(X')) + X',
\end{equation}
where $\text{LN}(\cdot)$ denotes Layer Normalization, and the proposed Superpixel-guided Selective Scan Mechanism ($\text{S}^3$M) module serves as the primary spatial mixer. 

The $\text{S}^3$M module integrates two primary components: \textit{Superpixel-guided Scanning} and the \textit{Region-level Gating Module}. Specifically, the scanning mechanism partitions the input into perceptually coherent regions via superpixel clustering and performs relations modeling within these semantic-related regions. This process constrains state propagation to prevent learning non-discriminative features from semantic-conflict areas. Simultaneously, the Region-level Gating Module is developed to perform intra-region calibration by modulating degradation outliers within each semantic superpixel unit along the channel dimension. The channel-wise details are subsequently processed by the Feed-Forward Network (FFN) to produce the final output $X_{out}$.

\subsubsection{Superpixel-guided Selective Scan Mechanism}
\label{sec:s3m_scanning}

The Superpixel-guided Scanning mechanism partitions the spatial domain of the intermediate feature $X_{mid} \in \mathbb{R}^{H \times W \times D}$ into $R$ perceptually coherent regions $\{\mathcal{S}_r\}_{r=1}^{R}$ using a superpixel generator~\cite{zhang2023lightweight}, where each pixel $(i,j)$ is assigned a region label $\ell_{i,j} \in \{1,\dots,R\}$. This partitioning ensures that the subsequent state propagation is confined within semantic-related regions to prevent learning non-discriminative features from semantic-conflict areas.

To align the state evolution with the semantic regions, we define a permutation operator $\mathcal{P}_{\ell}$ that reorders the flattened spatial tokens such that those belonging to the same superpixel are grouped contiguously. The superpixel-guided sequence modeling is formulated as:
\begin{equation}
    F_{seq}^{sp} = \mathcal{P}_{\ell}(\text{reshape}(X_{mid})), \quad Z_{seq}^{sp} = \text{SSM}(F_{seq}^{sp}),
\end{equation}
where $\text{SSM}(\cdot)$ denotes the selective scan operator. The output sequence is kemudian restored to the original spatial arrangement via an inverse permutation $Y = \text{reshape}(\mathcal{P}_{\ell}^{-1}(Z_{seq}^{sp}))$. This mechanism ensures that the state propagation is confined within perceptually coherent regions, thereby preventing the aggregation of non-discriminative features from semantic-conflict areas.

\subsubsection{Region-level Gating Mechanism }
\label{sec:rgm}

\begin{figure}[t!]
    \centering
    \includegraphics[width=0.75\linewidth]{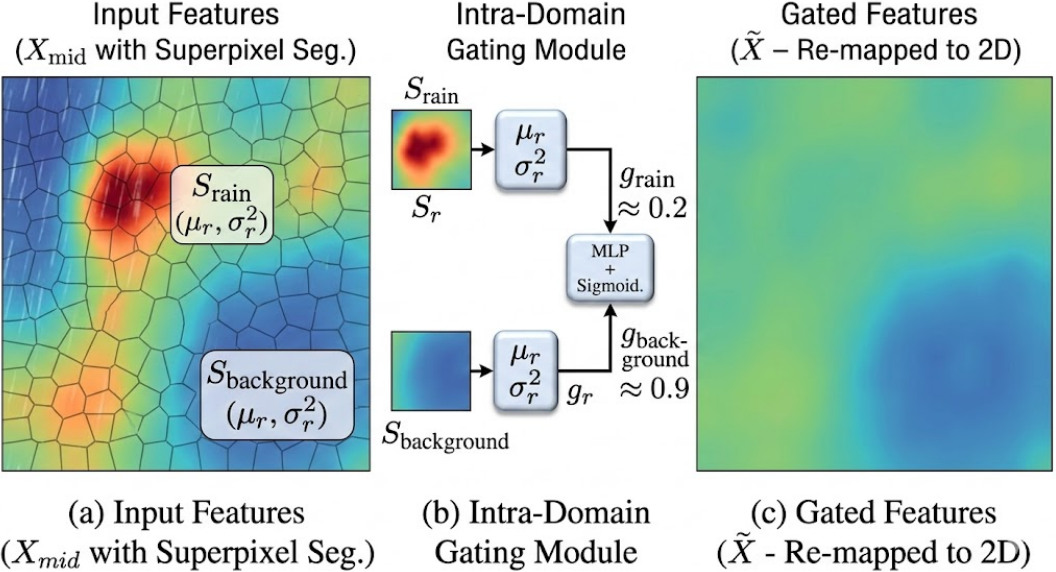}
    \vspace{-0.2cm}
    \caption{
        \textbf{Illustration of the Region-level Gating Mechanism (RGM).} 
        (a) Input features $X_{mid}$ with superpixel labels partitioning the spatial domain into perceptually coherent regions. 
        (b) The RGM computes regional statistics ($\mu_r, \sigma_r^2$) and employs a lightweight MLP to generate adaptive gating factors $g_r$ for channel-wise modulation. 
        (c) Visualized gated features $\tilde{X}$. The mechanism adaptively modulates degradation outliers within different semantic units to perform intra-region calibration.
    }
    \label{fig:gating_viz} 
\end{figure}

The Region-level Gating Mechanism (RGM) is developed to perform intra-region calibration by modulating degradation outliers along the channel dimension (Fig.~\ref{fig:gating_viz}). For each perceptually coherent region $\mathcal{S}_r$, we first compute the regional mean $\mu_r \in \mathbb{R}^C$ and variance $\sigma_r^2 \in \mathbb{R}^C$ from the intermediate feature $X_{mid}$ as follows:
\begin{equation}
    \mu_r = \frac{1}{|\mathcal{S}_r|} \sum_{(i,j)\in \mathcal{S}_r} X_{mid}^{(i,j)}, \quad \sigma_r^2 = \frac{1}{|\mathcal{S}_r|} \sum_{(i,j)\in \mathcal{S}_r} (X_{mid}^{(i,j)} - \mu_r)^2.
\end{equation}
To perceive the degradation characteristics within each semantic unit, these statistics are concatenated as $[\mu_r; \sigma_r^2] \in \mathbb{R}^{2C}$ and processed by a lightweight MLP. The MLP, comprising two fully-connected layers with a non-linear activation, generates a region-specific gating factor $g_r \in \mathbb{R}^C$. This factor adaptively modulates the channel-wise feature responses to perform intra-region calibration. For the $k$-th token $x_k \in F_{seq}^{sp}$ belonging to region $\mathcal{S}_r$, the gated update and subsequent state evolution are formulated as:
\begin{equation}
    \tilde{x}_k = \sigma(g_{r}) \odot x_k,
\end{equation}
\begin{equation}
    h_k = \bar{\mathbf{A}} h_{k-1} + \bar{\mathbf{B}} \tilde{x}_k,
\end{equation}
where $\sigma(\cdot)$ denotes the Sigmoid function, $\odot$ represents the element-wise product, and $h_k$ is the hidden state. By modulating degradation outliers within each semantic superpixel unit, the RGM suppresses non-discriminative features from semantic-conflict regions and provides a content-adaptive state transition.

\subsection{Loss Function}
\label{sec:loss_function}

To train the proposed SSR, we optimize the network using a combination of a pixel-wise reconstruction loss and a sequence-level structural correlation loss. 

Given the restored image $\hat{I}$ and the corresponding ground truth $I_{gt}$, we first employ the $L_1$ loss to constrain the fundamental pixel-level fidelity:
\begin{equation}
\mathcal{L}_{pixel} = \| \hat{I} - I_{gt} \|_1.
\end{equation}

While the $L_1$ loss ensures global color and luminance accuracy, relying solely on it can sometimes lead to overly smoothed results or geometric distortions in heavily degraded regions. To further enhance the structural coherence and linear correlation of the restored features, we introduce a Pearson correlation loss $\mathcal{L}_{seq}$. Let $x$ and $y$ denote the flattened 1D sequences of $\hat{I}$ and $I_{gt}$, respectively. The sequence-level Pearson loss is defined as:
\begin{equation}
\mathcal{L}_{seq} = 1 - \frac{\sum_{i} (x_i - \bar{x})(y_i - \bar{y})}{\sqrt{\sum_{i} (x_i - \bar{x})^2} \sqrt{\sum_{i} (y_i - \bar{y})^2}},
\end{equation}
where $\bar{x}$ and $\bar{y}$ are the mean values of the respective sequences. This term explicitly penalizes structural discrepancies and encourages the network to recover fine-grained details that are strictly correlated with the underlying clear image structure.

The overall objective function is the unweighted sum of these two terms:
\begin{equation}
\mathcal{L}_{total} = \mathcal{L}_{pixel} + \mathcal{L}_{seq}.
\end{equation}

\section{Experiments}

\begin{figure*}[t!]
    \centering
    \renewcommand{\arraystretch}{0}
    \setlength{\tabcolsep}{0.1pt} 

    \begin{tabular}{@{}cccccc@{}}
        % --- 第一行: Raindrop removal ---
        \includegraphics[width=0.165\textwidth]{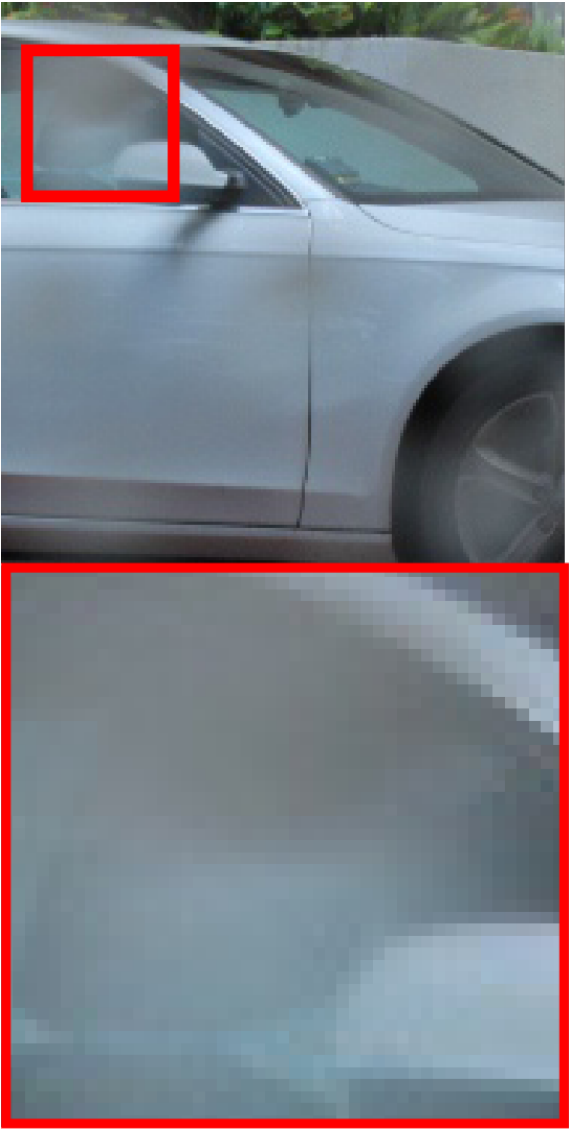} &
        \includegraphics[width=0.165\textwidth]{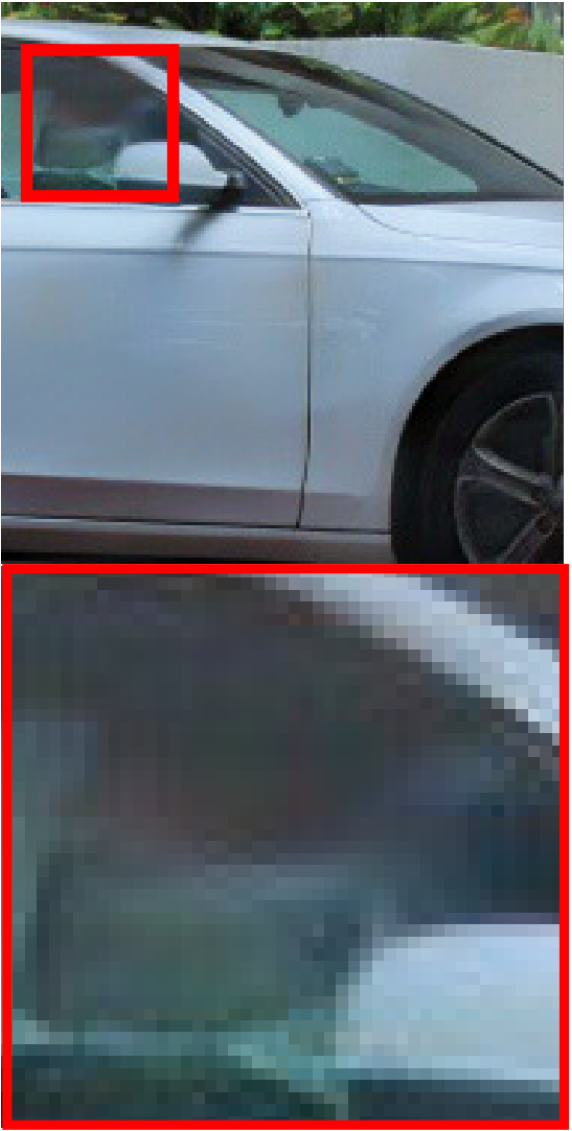} &
        \includegraphics[width=0.165\textwidth]{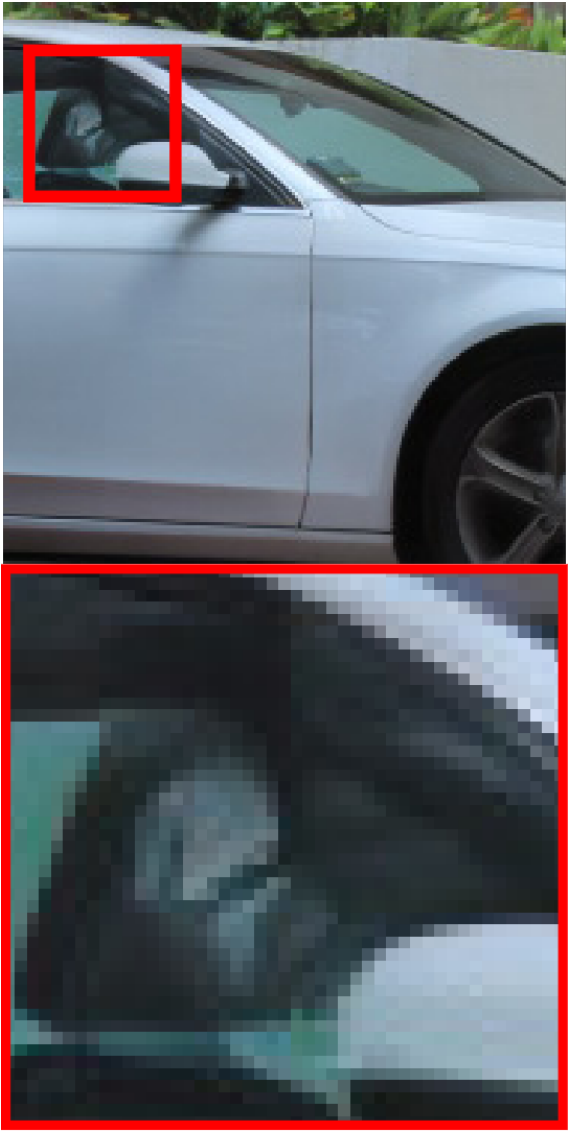} &
        \includegraphics[width=0.165\textwidth]{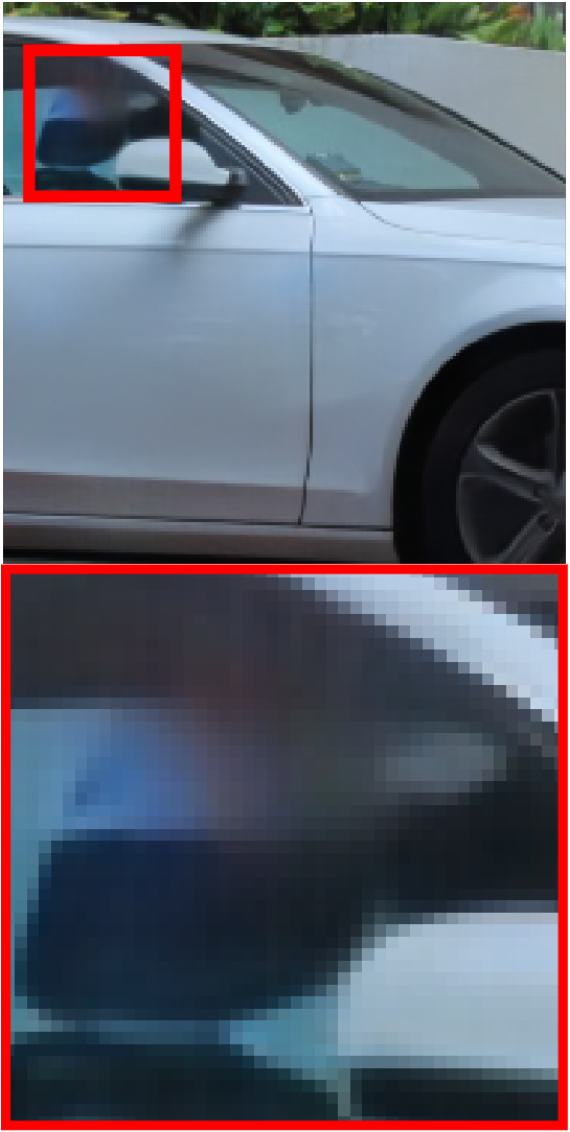} &
        \includegraphics[width=0.165\textwidth]{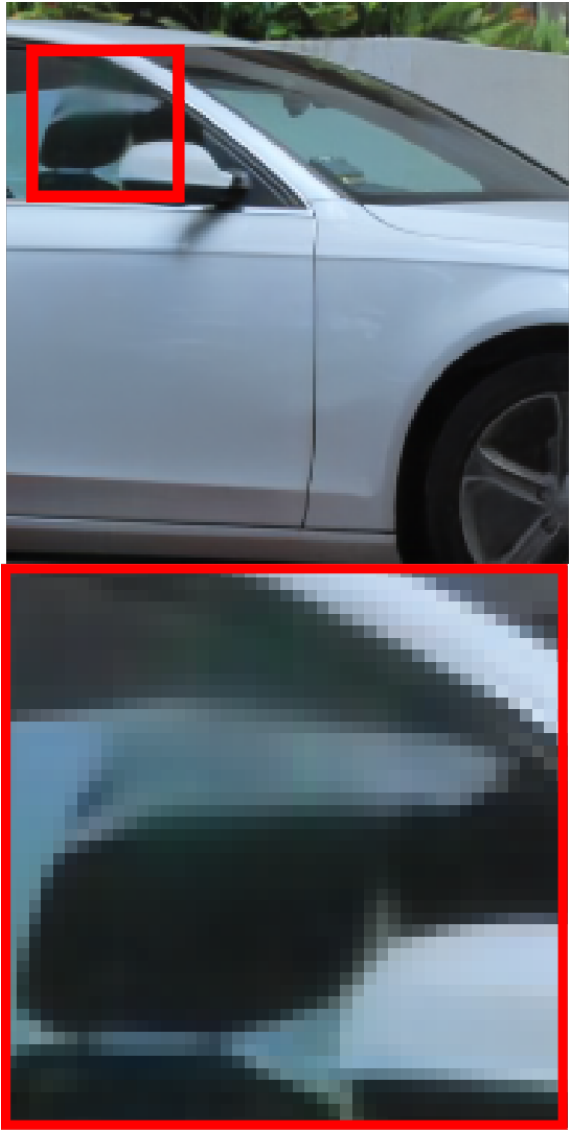} &
        \includegraphics[width=0.165\textwidth]{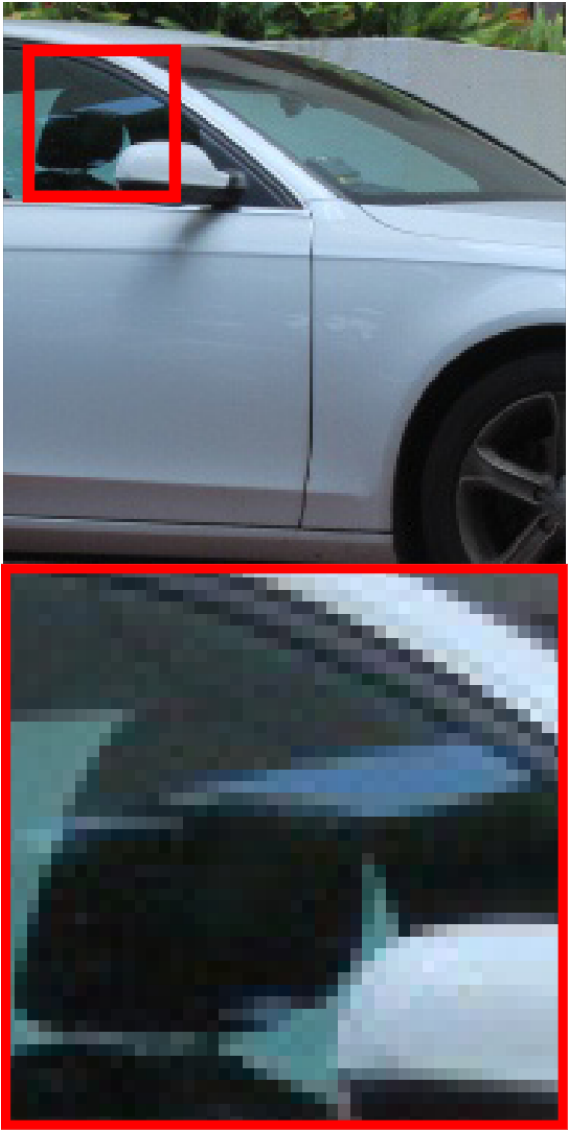} \\
        
        % --- 第二行: Joint deraining and dehazing ---
        \includegraphics[width=0.165\textwidth]{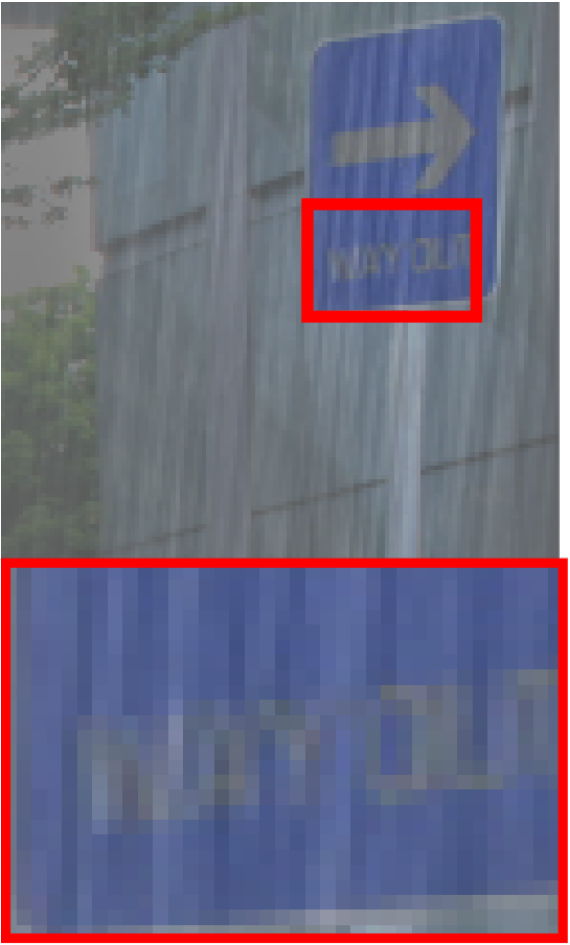} &
        \includegraphics[width=0.165\textwidth]{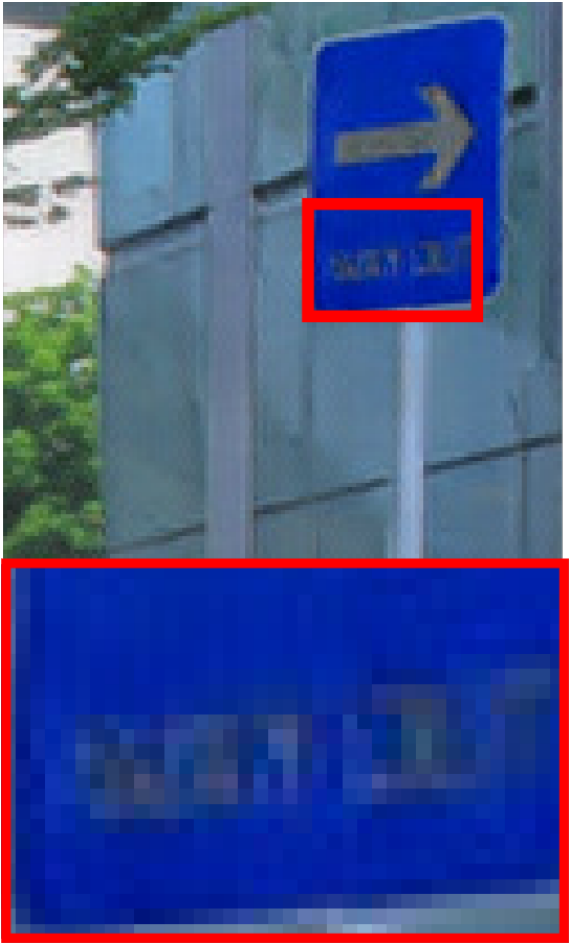} &
        \includegraphics[width=0.165\textwidth]{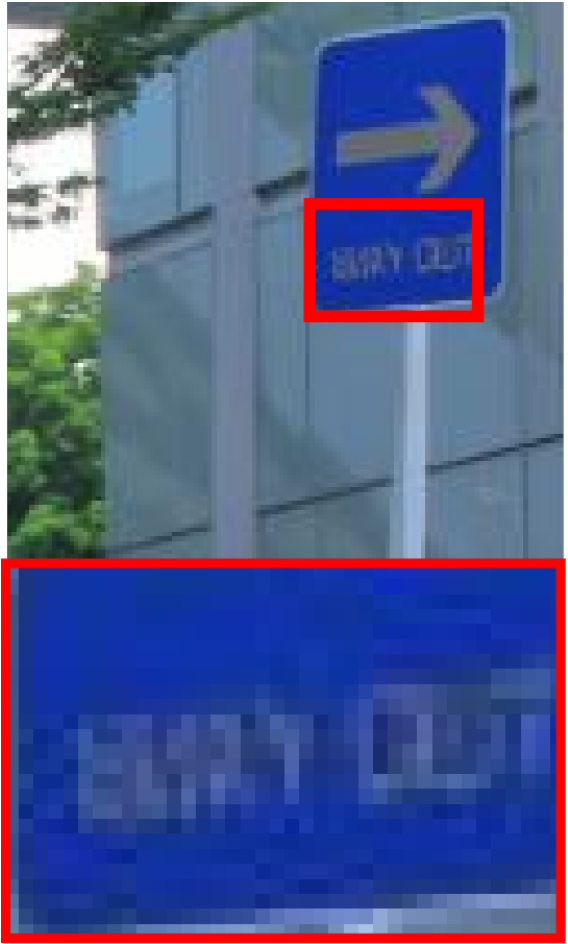} &
        \includegraphics[width=0.165\textwidth]{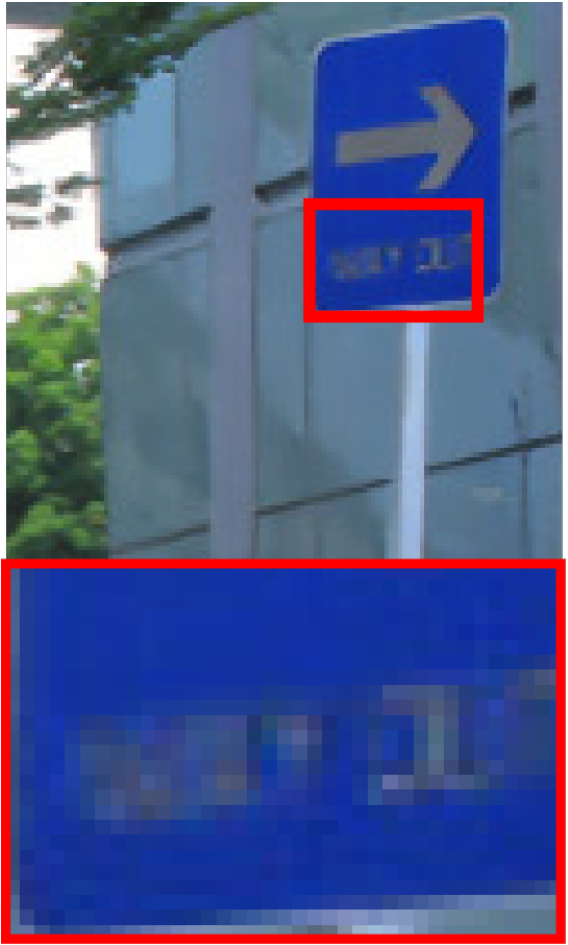} &
        \includegraphics[width=0.165\textwidth]{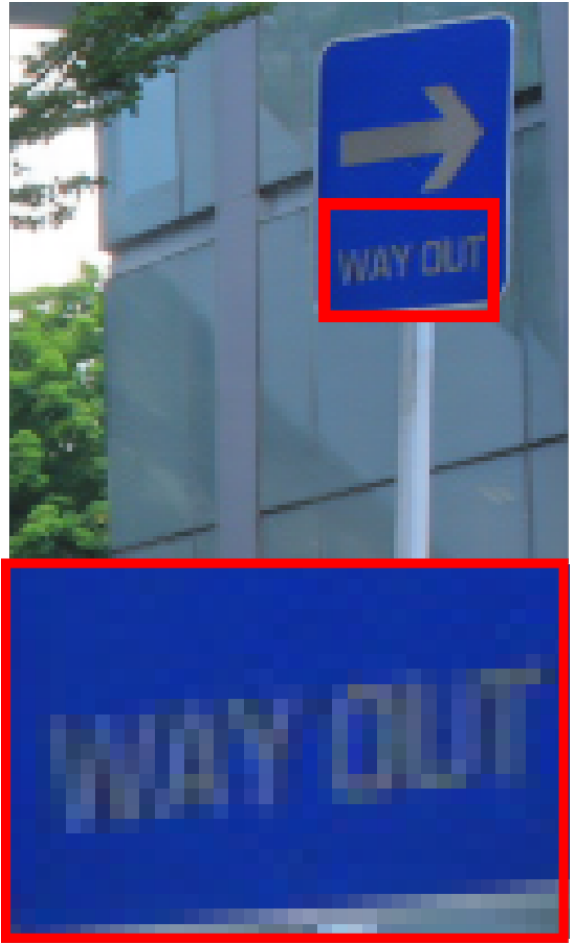} &
        \includegraphics[width=0.165\textwidth]{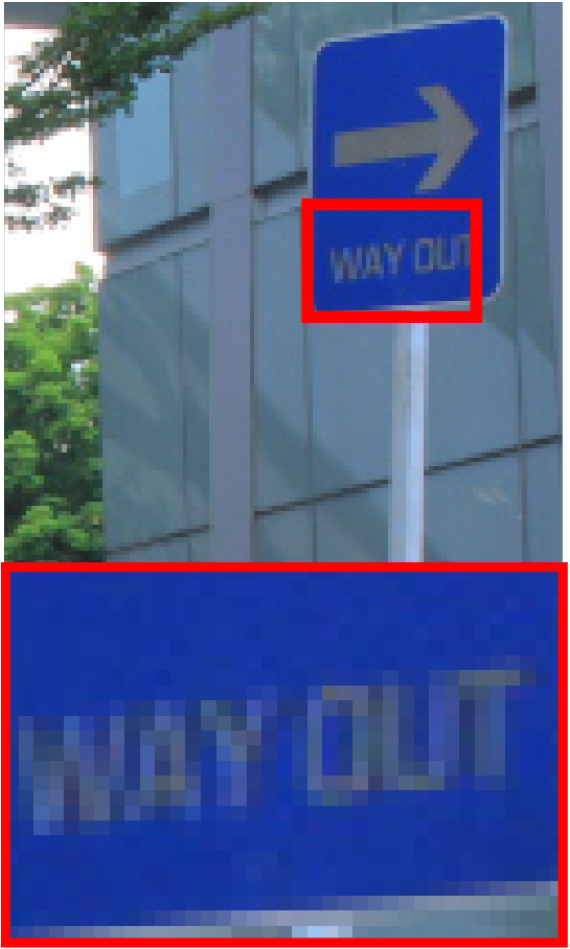} \\
        
        % --- 第三行: Image desnowing ---
        \includegraphics[width=0.165\textwidth]{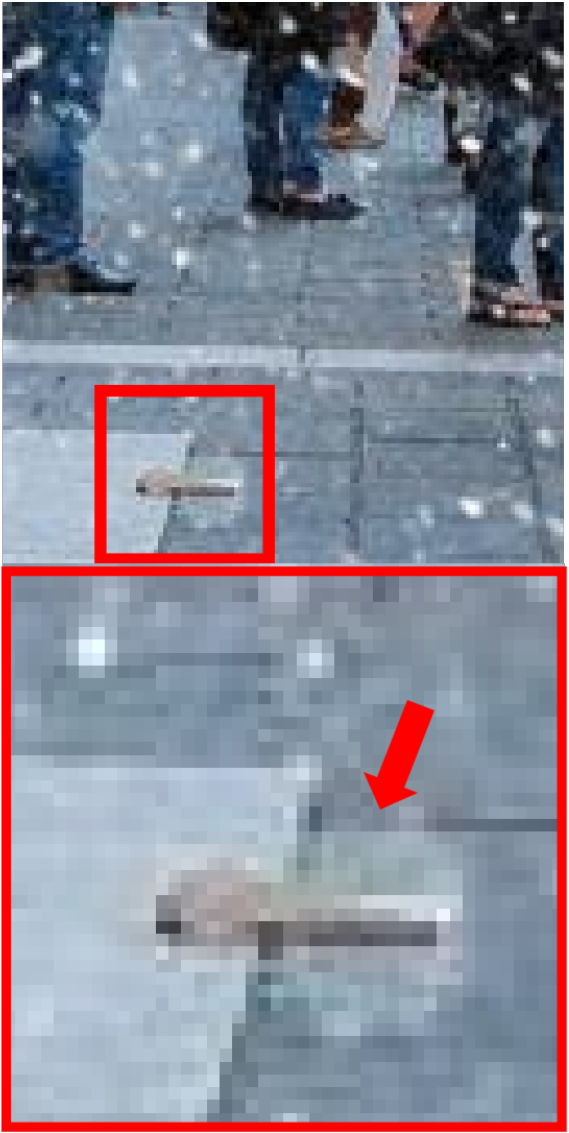} &
        \includegraphics[width=0.165\textwidth]{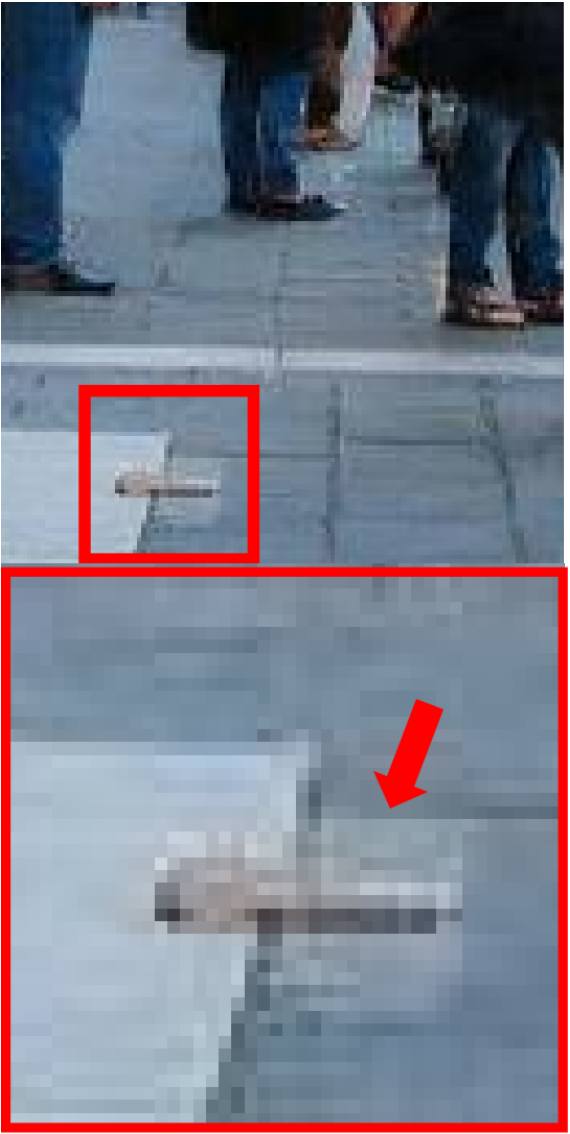} &
        \includegraphics[width=0.165\textwidth]{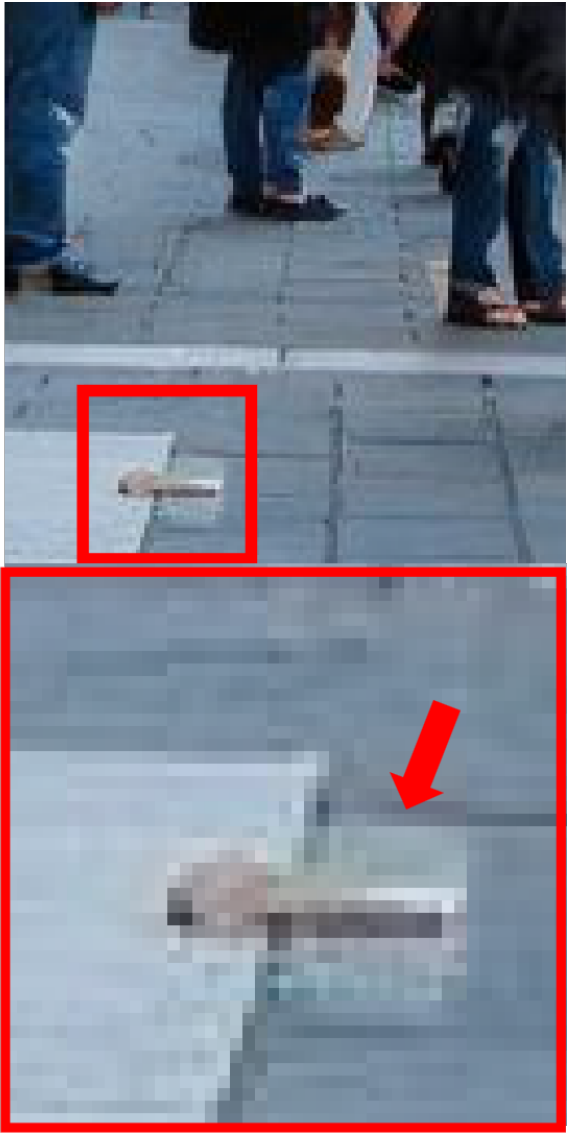} &
        \includegraphics[width=0.165\textwidth]{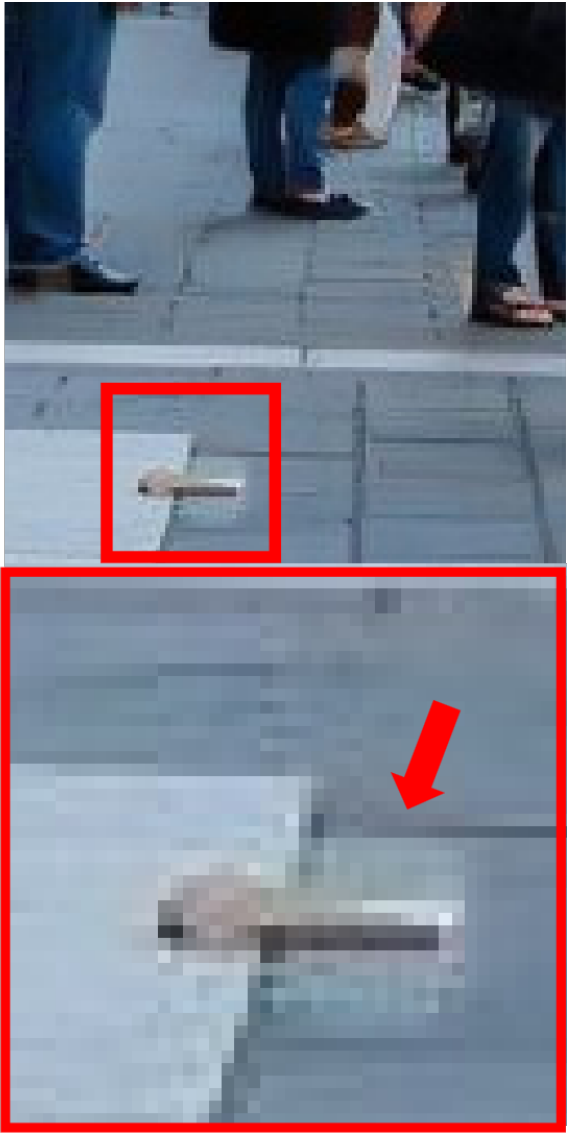} &
        \includegraphics[width=0.165\textwidth]{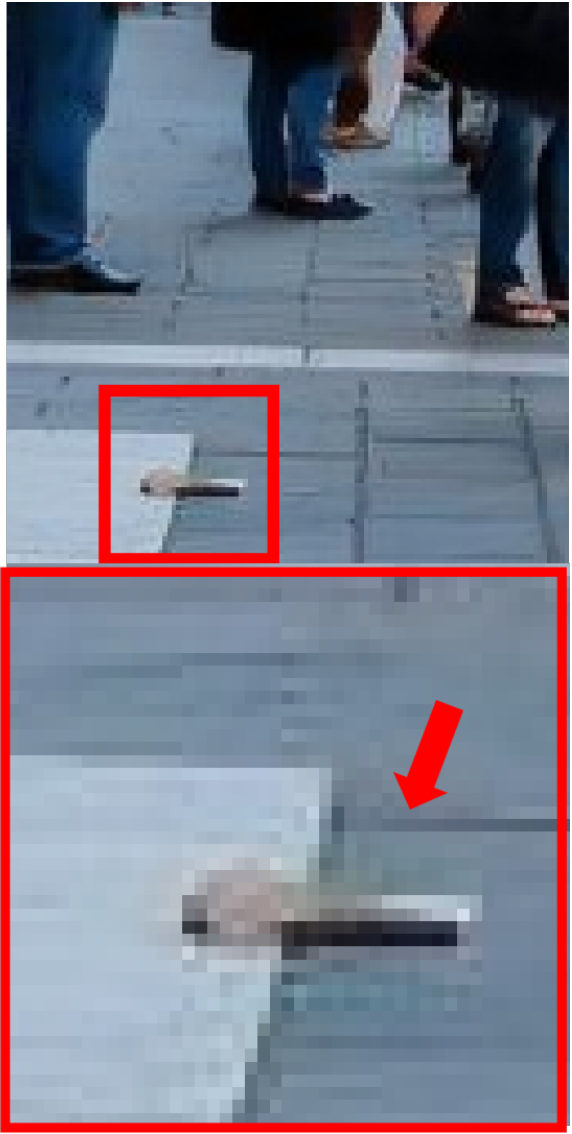} &
        \includegraphics[width=0.165\textwidth]{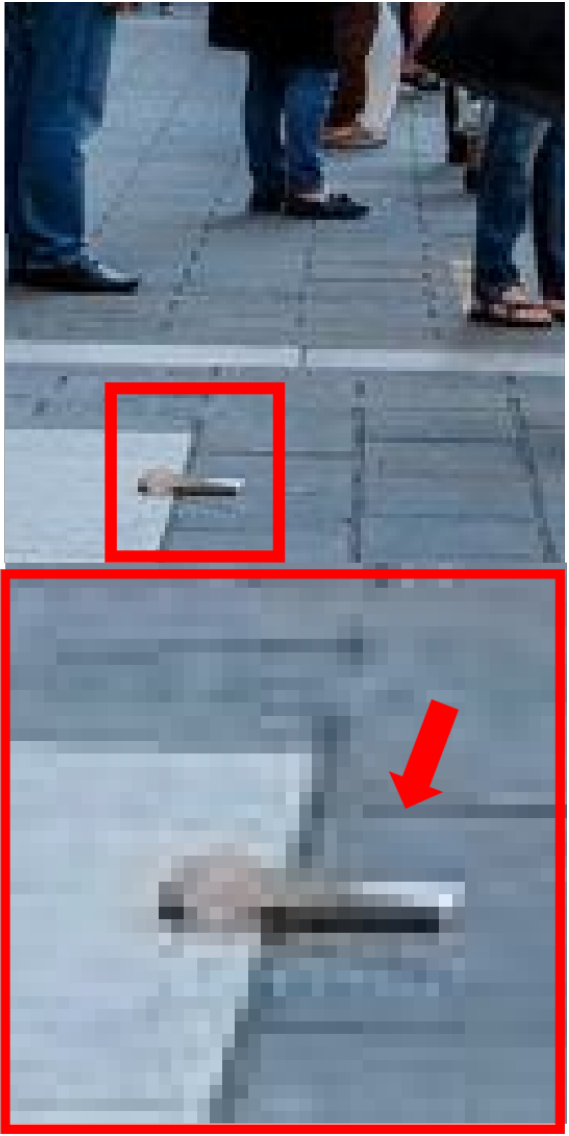} \\

        \noalign{\smallskip}
        \scriptsize (a)Input & \scriptsize (b)TransW. & \scriptsize (c)WeatherDiff & 
        \scriptsize (d)Histoformer & \scriptsize (e)Ours & \scriptsize (f)GT
    \end{tabular}

    \vspace{-2mm}
    \caption{
        Visual comparisons of SSR against state-of-the-art unified methods. \textbf{Top Row}: Raindrop removal~\cite{ozdenizci2023restoring}. \textbf{Middle Row}: Joint deraining and dehazing~\cite{li2019heavy,li2020all}. \textbf{Bottom Row}: Image desnowing~\cite{liu2018desnownet}.
    }
    \label{fig:com}
\end{figure*}

\subsection{Experiment Settings}

\subsubsection{Datasets and Evaluation Metrics.}
To ensure fair comparison, we follow the standard experimental protocol adopted by previous all-in-one adverse weather restoration methods~\cite{li2020all,ozdenizci2023restoring,valanarasu2022transweather,sun2024restoring}.
The training set consists of 9,000 images from Snow100K~\cite{liu2018desnownet}, 1,069 images from Raindrop~\cite{qian2018attentive}, and 9,000 images from Outdoor-Rain~\cite{li2019heavy}.
Snow100K and Outdoor-Rain are synthetic datasets containing snow, rain, and fog degradations, while Raindrop comprises real-world images degraded by raindrops.
For evaluation, we adopt the Test1 set~\cite{li2019heavy,li2020all}, the Raindrop test set~\cite{qian2018attentive}, and the Snow100K-S and Snow100K-L test sets~\cite{liu2018desnownet}.
In addition, Snow100K provides a real-world test set containing 1,329 images with real adverse weather degradations.
We report Peak Signal-to-Noise Ratio (PSNR) and Structural Similarity Index (SSIM) as quantitative evaluation metrics, following common practice in adverse weather image restoration\cite{li2020all}.

\subsubsection{Implementation Details.}
Our model is implemented in PyTorch~\cite{paszke2019pytorch} and trained on a single NVIDIA A40 GPU.
The network is trained for 300,000 iterations with an initial batch size of 8.
Following the progressive learning strategy in Restormer~\cite{zamir2022restormer}, we use an initial patch size of 128.
We employ the AdamW optimizer~\cite{loshchilov2019decoupled} with an initial learning rate of $3 \times 10^{-4}$ for the first 92,000 iterations, which is then gradually decayed to $1 \times 10^{-6}$ using a cosine annealing schedule~\cite{loshchilov2017sgdr} over the remaining iterations.
The number of blocks at each stage is set to $N_{1,2,3}=\{3,3,6\}$, the base channel dimension is 48, and the channel expansion ratio is set to 3.
Standard data augmentation including random horizontal and vertical flipping is applied during training.

\begin{table}[tb]
  \caption{Quantitative comparisons of adverse weather restoration. The \textbf{best} and \underline{second-best} results are highlighted.}
  \label{tab:comparison}
  \centering
  \setlength{\aboverulesep}{0pt}
  \setlength{\belowrulesep}{0pt}
  \renewcommand{\arraystretch}{1.2} 
  
  \resizebox{\textwidth}{!}{
    \scriptsize  
    \setlength{\tabcolsep}{3pt} 
    \begin{tabular}{@{\hspace{2pt}} l | *{8}{c} | *{2}{c} @{\hspace{2pt}}}
      \Xhline{1.2pt} 
      \noalign{\vspace{2pt}}
      \multirow{2}{*}{Methods} & \multicolumn{2}{c}{Snow100K-S} & \multicolumn{2}{c}{Snow100K-L} & \multicolumn{2}{c}{Outdoor} & \multicolumn{2}{c|}{RainDrop} & \multicolumn{2}{c}{Average} \\
      \cmidrule(lr){2-3} \cmidrule(lr){4-5} \cmidrule(lr){6-7} \cmidrule(lr){8-9} \cmidrule(lr){10-11}
      & PSNR & SSIM & PSNR & SSIM & PSNR & SSIM & PSNR & \multicolumn{1}{c|}{SSIM} & PSNR & SSIM \\
      \Xhline{0.8pt} 
      % 2020
      All-in-One~\cite{li2020all} \tiny{(CVPR'20)} & - & - & 28.33 & 0.8820 & 24.71 & 0.8980 & 31.12 & 0.9268 & - & - \\
      % 2021
      SwinIR~\cite{liang2021swinir} \tiny{(ICCV'21)} & - & - & 28.18 & 0.880 & 23.23 & 0.869 & 30.82 & 0.904 & - & - \\
      MPRNet~\cite{zamir2021multi} \tiny{(CVPR'21)} & - & - & 28.66 & 0.869 & 30.25 & 0.914 & 30.99 & 0.916 & - & - \\
      TKMANet~\cite{wang2021temporal} \tiny{(ICIP'21)} & - & - & 30.24 & 0.902 & 29.92 & 0.917 & 30.99 & 0.927 & - & - \\
      % 2022
      AirNet~\cite{li2022all} \tiny{(CVPR'22)} & - & - & 27.92 & 0.858 & 23.12 & 0.837 & 28.23 & 0.892 & - & - \\
      TransWeather~\cite{valanarasu2022transweather} \tiny{(CVPR'22)} & 32.51 & 0.9341 & 29.31 & 0.8879 & 28.83 & 0.9000 & 30.17 & 0.9157 & 30.21 & 0.9094 \\
      Chen et al.~\cite{chen2022learning} \tiny{(CVPR'22)} & 34.42 & 0.9469 & 30.22 & 0.9071 & 29.27 & 0.9147 & 31.81 & 0.9309 & 31.43 & 0.9249 \\
      Restormer~\cite{zamir2022restormer} \tiny{(CVPR'22)} & 36.01 & 0.9579 & 30.36 & 0.9068 & 30.03 & 0.9215 & 32.18 & 0.9408 & 32.15 & 0.9318 \\
      % 2023
      WeatherDiff$_{128}$~\cite{ozdenizci2023restoring} \tiny{(TPAMI'23)} & 35.02 & 0.9652 & 29.58 & 0.8941 & 29.72 & 0.9216 & 29.66 & 0.9225 & 31.00 & 0.9225 \\
      WGWSNet~\cite{zhu2023learning} \tiny{(CVPR'23)} & 34.31 & 0.9460 & 30.16 & 0.9007 & 29.32 & 0.9207 & 32.38 & 0.9378 & 31.54 & 0.9263 \\
      WeatherDiff$_{64}$~\cite{ozdenizci2023restoring} \tiny{(TPAMI'23)} & 35.83 & 0.9566 & 30.09 & 0.9041 & 29.64 & 0.9312 & 30.71 & 0.9312 & 31.57 & 0.9308 \\
      AWRCP~\cite{ye2023adverse} \tiny{(ICCV'23)} & 36.92 & 0.9652 & 31.92 & \textbf{0.9341} & 31.39 & 0.9329 & 31.93 & 0.9314 & 33.04 & 0.9409 \\
      % 2024
      GridFormer~\cite{wang2024gridformer} \tiny{(IJCV'24)} & 37.46 & 0.9640 & 31.71 & 0.9231 & 31.87 & 0.9335 & 32.39 & 0.9362 & 33.36 & 0.9392 \\
      Histoformer~\cite{peng2024histoformer} \tiny{(ECCV'24)} & 37.41 & 0.9656 & 32.16 & 0.9261 & 32.08 & 0.9389 & \textbf{33.06} & \textbf{0.9441} & 33.68 & 0.9437 \\
      % 2025
      Mambairv2~\cite{guo2025mambairv2} \tiny{(CVPR'25)} & 37.13 & 0.9655 & 31.75 & 0.9234 & 31.45 & 0.9332 & 32.22 & 0.9373 & 33.14 & 0.9398 \\
      EVSSM~\cite{kong2025efficient} \tiny{(CVPR'25)} & 37.45 & 0.9655 & 32.15 & 0.9250 & 32.06 & 0.9380 & 32.40 & 0.9400 & 33.52 & 0.9421 \\
      MoCE-IR~\cite{zamfir2025complexity} \tiny{(CVPR'25)} & 37.10 & 0.9654 & 31.88 & 0.9234 & 31.42 & 0.9334 & 32.23 & 0.9386 & 33.16 & 0.9400 \\
      % 2026
      CPL$_{PromptIR}$~\cite{wu2025beyond} \tiny{(TPAMI'26)} & \underline{37.82} & \textbf{0.9667} & \underline{32.27} & 0.9280 & \underline{32.16} & \underline{0.9420} & 32.73 & 0.9430 & \underline{33.74} & \underline{0.9438} \\
      \hline
      SSR\tiny{(Ours)} & \textbf{37.87} & \underline{0.9656} & \textbf{32.70} & \underline{0.9297} & \textbf{33.22} & \textbf{0.9421} & \underline{32.82} & \underline{0.9440} & \textbf{34.15} & \textbf{0.9442} \\
      \Xhline{1.2pt} 
    \end{tabular}
  }
\end{table}

\subsection{Comparison with State-of-the-Art Methods}
\label{sec:experiments}

\subsubsection{Quantitative Comparison.} 
We evaluate the performance of SSR against a diverse set of state-of-the-art (SOTA) image restoration methods. These include Transformer-based models such as Restormer~\cite{zamir2022restormer}, AWRCP~\cite{ye2023adverse}, and Histoformer~\cite{peng2024histoformer}, as well as contemporary SSM-based frameworks like MambaIRv2~\cite{guo2025mambairv2} and EVSSM~\cite{kong2025efficient}.
As reported in \cref{tab:comparison}, SSR achieves SOTA performance across most benchmarks. Notably, on the challenging Outdoor dataset, which consists of complex joint rain and haze degradations, our method outperforms the second-best competitor (CPL$_{PromptIR}$~\cite{wu2025beyond}) by a significant margin of 1.06 dB in PSNR. Furthermore, SSR attains the best results in terms of average PSNR (34.15 dB) and SSIM (0.9442) across all test sets, demonstrating its superior robustness and generalization capability in handling spatially non-uniform degradations. 
While Histoformer~\cite{peng2024histoformer} and CPL$_{PromptIR}$~\cite{wu2025beyond} achieve slightly better results on specific synthetic benchmarks (e.g., Snow100K-S and RainDrop), SSR consistently ranks as the best or second-best method across these tasks. This confirms that our proposed superpixel-guided strategy effectively handles localized artifacts (like raindrops) without sacrificing the modeling of large-scale structural context.

\subsubsection{Qualitative Comparison}
We provide visual comparisons of SSR against leading unified methods across three diverse adverse weather scenarios in Fig.~\ref{fig:com} 
On the raindrop removal task (Top Row), while methods such as TransWeather~\cite{valanarasu2022transweather} (b) and WeatherDiff~\cite{ozdenizci2023restoring} (c) struggle to restore the car's interior details obscured by heavy drops, SSR effectively eliminates the artifacts while preserving sharp structural content. For the challenging joint deraining and dehazing task (Middle Row), SSR excels in restoring fine texture details; notably, the text "WAY OUT" on the blue sign remains clearly legible in our result, validating the efficacy of our semantic-guided scanning mechanism. Finally, in the image desnowing task (Bottom Row), SSR demonstrates superior detail preservation by eliminating dense snowflakes without introducing blurring artifacts, especially on the sharp edges of small objects (indicated by red arrows). 
Overall, these visual results underscore SSR's remarkable capability to adaptively model spatially non-uniform weather degradations while maintaining high perceptual fidelity.

\subsubsection{Efficiency and Model Complexity Analysis}
Based on the data presented in \cref{tab:comparison_effection}, we comprehensively evaluate the proposed SSR from the perspectives of performance, parameter efficiency, and computational complexity. Compared to a broad range of state-of-the-art unified restoration models, SSR achieves superior performance on the Outdoor dataset (33.22 dB) while maintaining remarkable parameter efficiency. As shown in \cref{tab:comparison_effection}, our method utilizes only 7.05M parameters, which is the lowest among all compared approaches. 

\begin{wraptable}{r}{0.45\textwidth}
  \vspace{-2.2em}
  \caption{Performance and Complexity Comparison on the Outdoor Dataset.}
  \label{tab:comparison_effection}
  \centering
  \scriptsize
  \resizebox{0.45\textwidth}{!}{
    \setlength{\tabcolsep}{4pt} 
    \begin{tabular}{l c c c}
      \toprule
      Method & Param./M & FLOPs/G & PSNR/dB \\
      \midrule
      AirNet~\cite{li2022all} & \underline{8.93} & \textbf{30.13} & 23.12 \\
      Restormer~\cite{zamir2022restormer} & 26.12 & 141.00 & 30.03 \\
      WeatherDiff~\cite{ozdenizci2023restoring} & 82.96 & -- & 29.64 \\
      Histoformer~\cite{peng2024histoformer} & 16.61 & 91.65 & 32.08 \\
      MambairV2~\cite{guo2025mambairv2} & 15.42 & 76.50 & 31.45 \\
      EVSSM~\cite{kong2025efficient} & 17.13 & 122.23 & 32.06 \\
      CPL$_{PromptIR}$~\cite{wu2025beyond} & 36.00 & -- & 32.16 \\
      \midrule
      \textbf{SSR (Ours)} & \textbf{7.05} & \underline{54.27} & \textbf{33.22} \\
      \bottomrule
    \end{tabular}
  }
  \vspace{-1.5em}
\end{wraptable}

Compared to recent high-performing methods such as CPL$_{PromptIR}$~\cite{wu2025beyond}, SSR achieves a significant performance gain of +1.06 dB while reducing the model size by approximately 80.4\%. Even against the second most parameter-efficient model, AirNet~\cite{li2022all}, our method demonstrates a clear advantage in both restoration quality (+10.10 dB) and parameter economy (-1.88M).

In terms of computational complexity measured by FLOPs, SSR strikes a highly favorable balance between restoration capability and execution efficiency. With 54.27G FLOPs, SSR requires less than half the computation of previous Transformer-based heavyweights like Restormer~\cite{zamir2022restormer} (141.00G) and is significantly more economical than recent unified models such as EVSSM~\cite{kong2025efficient} (122.23G) and Histoformer~\cite{peng2024histoformer} (91.65G). It also notably undercuts the baseline MambaIRv2~\cite{guo2025mambairv2} (76.50G). While AirNet~\cite{li2022all} achieves lower FLOPs (30.13G), it suffers from a severe performance degradation (23.12 dB) compared to SSR. Overall, these quantitative results rigorously validate that our proposed SSR is a highly competitive unified weather restoration model, achieving SOTA performance while maintaining high efficiency in both model footprint and computational cost.

\subsubsection{Real-World Evaluation}

\begin{table}[h]
  \caption{Quantitative comparison on real-world datasets. Left: Evaluation using Q-Align ($\uparrow$). Right: Evaluation using NIQE and IL-NIQE ($\downarrow$) along with parameter comparisons.}
  \label{tab:realworld_comprehensive}
  \centering
  \begin{minipage}{0.46\textwidth}
    \centering
    \resizebox{\linewidth}{!}{
    \setlength{\tabcolsep}{3pt}
    \begin{tabular}{l c c c c}
      \toprule
      \multirow{2}{*}{Method} & \multicolumn{2}{c}{RainDrop-real} & \multicolumn{2}{c}{NTURain-real} \\
      \cmidrule(lr){2-3} \cmidrule(lr){4-5} 
      & Quality & Aesthetic & Quality & Aesthetic \\
      \midrule
      WeatherDiff~\cite{ozdenizci2023restoring} & 0.6550 & 0.3231 & 0.6582 & 0.2537 \\
      Histoformer~\cite{peng2024histoformer}   & 0.6551 & 0.3234 & 0.6584 & 0.2541 \\
      \textbf{SSR (Ours)}                      & \textbf{0.6591} & \textbf{0.3246} & \textbf{0.6587} & \textbf{0.2561} \\
      \bottomrule
    \end{tabular}
    }
  \end{minipage}
  \hfill
  \begin{minipage}{0.52\textwidth}
    \centering
    \resizebox{\linewidth}{!}{
    \setlength{\tabcolsep}{4pt}
    \begin{tabular}{l c c c c c}
      \toprule
      \multirow{2}{*}{Method} & \multicolumn{2}{c}{Snow100K-Real} & \multicolumn{2}{c}{RainDS} & \multirow{2}{*}{Param} \\
      \cmidrule(lr){2-3} \cmidrule(lr){4-5} 
      & NIQE$\downarrow$ & IL-NIQE$\downarrow$ & NIQE$\downarrow$ & IL-NIQE$\downarrow$ & \\
      \midrule
      TransWeather~\cite{valanarasu2022transweather} & 3.161 & 22.207 & 4.005 & 22.512 & 38.05M \\
      WeatherDiff\_64~\cite{ozdenizci2023restoring} & 2.985 & 22.121 & 3.050 & 19.800 & 82.96M \\
      WeatherDiff\_128~\cite{ozdenizci2023restoring} & 2.964 & 21.976 & 3.642 & 19.972 & 85.61M \\
      \textbf{SSR (Ours)} & \textbf{2.879} & \textbf{21.909} & \textbf{2.924} & \textbf{19.215} & \textbf{7.05M} \\
      \bottomrule
    \end{tabular}
    }
  \end{minipage}
\end{table}

\label{sec:realworld}
To provide quantitative evidence of our model's performance in challenging real-world scenarios, we evaluate our SSR framework on real-world adverse weather images. Since real-world degradations lack ground-truth clear images for standard reference-based evaluation (e.g., PSNR/SSIM), we employ no-reference metrics to evaluate the restored results. Specifically, we use the multi-modal no-reference metric, Q-Align~\cite{wu2023q}, to assess both perceptual \textit{quality} and \textit{aesthetic} scores on the real-world subsets of the RainDrop~\cite{ozdenizci2023restoring} and NTURain~\cite{CVPR2021_2429} datasets. In addition, we further evaluate our model on real-world datasets by adopting NIQE and IL-NIQE on Snow100K-Real and RainDS benchmarks. The results are presented in \cref{tab:realworld_comprehensive}.

As the results show, our method achieves state-of-the-art performance across all datasets. In the Q-Align dimensions, our method outperforms the recent Transformer-based method, Histoformer~\cite{peng2024histoformer}, and the diffusion-based approach, WeatherDiff~\cite{ozdenizci2023restoring}, in both quality and aesthetic scores. Meanwhile, on the Snow100K-Real and RainDS benchmarks, SSR outperforms existing methods by a clear margin while maintaining a significantly lower parameter budget.

\subsection{Ablation Study}

We conduct comprehensive ablation studies to analyze the impact of key design
choices in our framework.
Unless otherwise specified, all ablation experiments are performed under the
same SSM-based architecture, and only the corresponding scanning or grouping
strategy is modified.

\begin{table*}[h]
  \centering
  \caption{Ablation study on the number of superpixels ($K$) across different datasets.}
  \label{tab:ablation_k}
  \setlength{\aboverulesep}{0pt}
  \setlength{\belowrulesep}{0pt}
  \renewcommand{\arraystretch}{1.3}
  \resizebox{\textwidth}{!}{
    \footnotesize
    \begin{tabular}{@{\hspace{25pt}} c @{\hspace{25pt}} | c | *{8}{c} | *{2}{c}}
      \Xhline{1.2pt}
      \noalign{\vspace{2pt}}
      \multirow{2}{*}{\textbf{$K$}}
      & \multicolumn{1}{c|}{\multirow{2}{*}{FLOPs}}
      & \multicolumn{2}{c}{Snow100K-S}
      & \multicolumn{2}{c}{Snow100K-L}
      & \multicolumn{2}{c}{Outdoor}
      & \multicolumn{2}{c|}{RainDrop}
      & \multicolumn{2}{c}{Average} \\
      \cmidrule(lr){3-4} \cmidrule(lr){5-6} \cmidrule(lr){7-8} \cmidrule(lr){9-10} \cmidrule(lr){11-12}
      & \multicolumn{1}{c|}{}
      & PSNR & SSIM
      & PSNR & SSIM
      & PSNR & SSIM
      & PSNR & \multicolumn{1}{c|}{SSIM}
      & PSNR & SSIM \\
      \Xhline{0.8pt}
      8  & \textbf{4.21G} & 37.34 & 0.9715 & 31.62 & 0.9213 & 31.68 & 0.9218 & 32.25 & 0.9391 & 33.22 & 0.9384 \\
      16 & 4.30G          & \textbf{37.42} & \textbf{0.9717} & \textbf{31.90} & \textbf{0.9227} & \textbf{31.91} & \textbf{0.9239} & \textbf{32.37} & \textbf{0.9396} & \textbf{33.40} & \textbf{0.9395} \\
      32 & 4.71G          & 37.26 & 0.9714 & 31.88 & 0.9224 & 31.57 & 0.9212 & 32.28 & 0.9394 & 33.25 & 0.9386 \\
      \noalign{\vspace{2pt}}
      \Xhline{1.2pt}
    \end{tabular}
  }
\end{table*}

\subsubsection{Effect of the Number of Superpixels $K$.}

\begin{figure}[h!]
    \centering
    \begin{minipage}{0.24\linewidth}
        \centering
        \includegraphics[width=\linewidth]{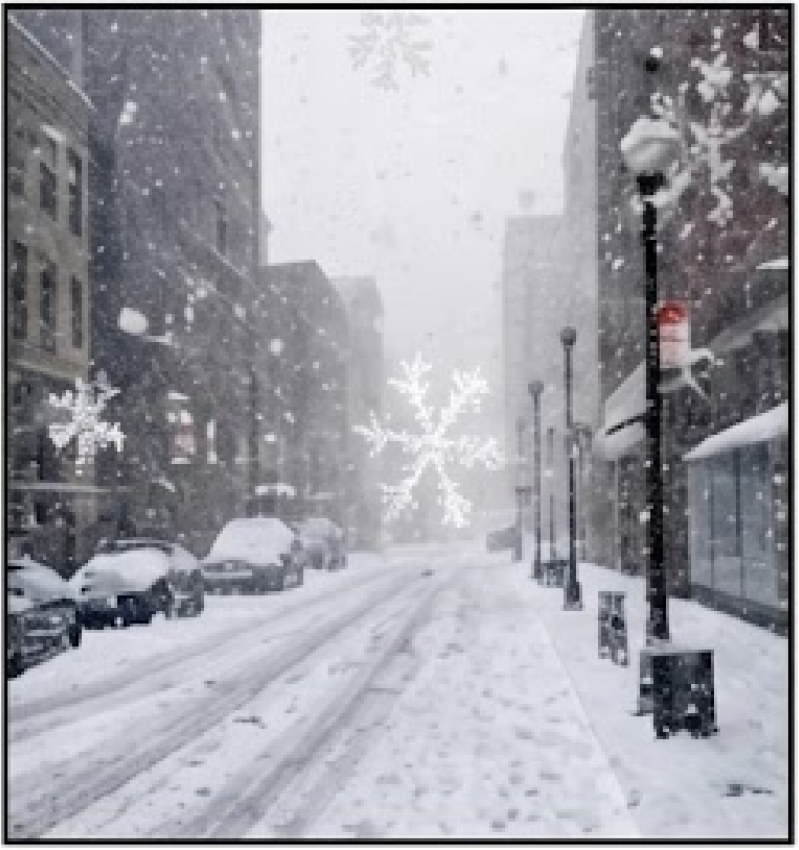}
        \\ \vspace{1mm}
        {\small (a) Input}
    \end{minipage}\hfill
    \begin{minipage}{0.24\linewidth}
        \centering
        \includegraphics[width=\linewidth]{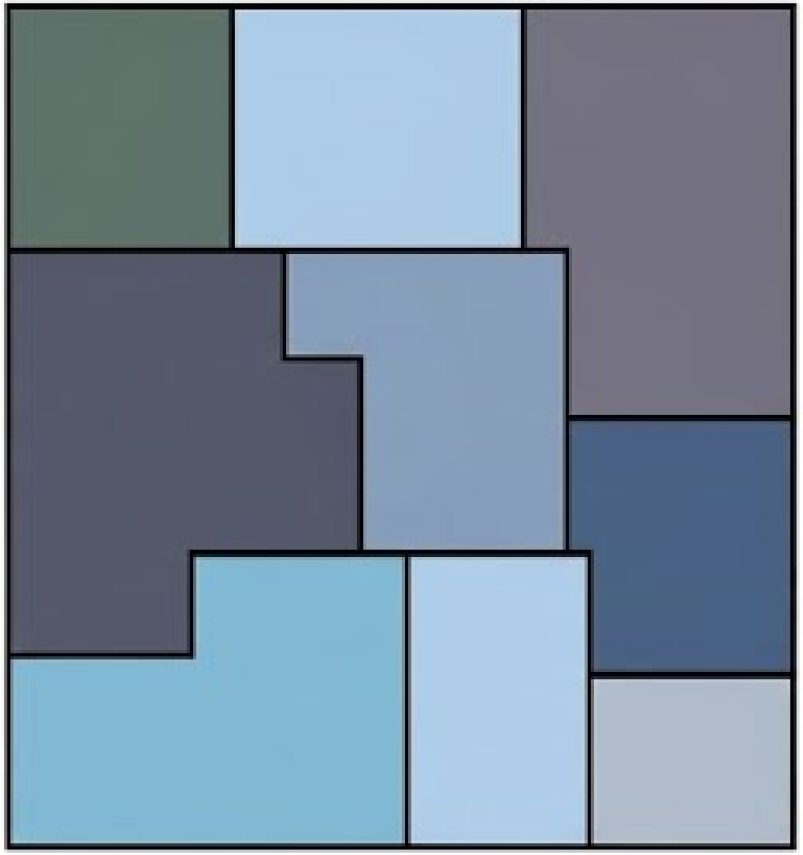}
        \\ \vspace{1mm}
        {\small (b) $K=8$}
    \end{minipage}\hfill
    \begin{minipage}{0.24\linewidth}
        \centering
        \includegraphics[width=\linewidth]{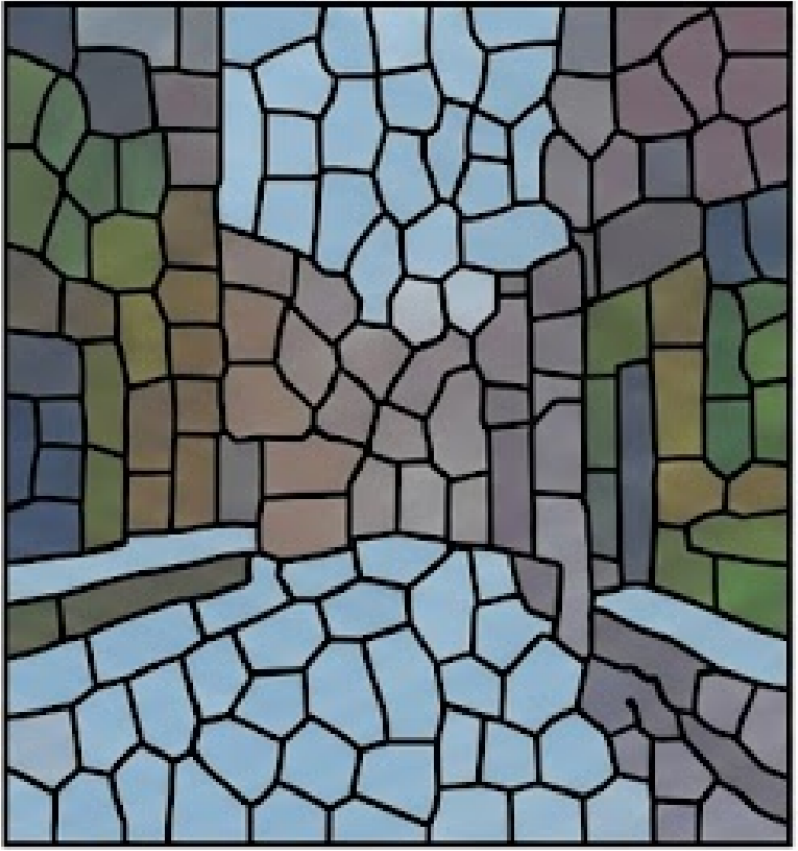}
        \\ \vspace{1mm}
        {\small (c) $K=16$}
    \end{minipage}\hfill
    \begin{minipage}{0.24\linewidth}
        \centering
        \includegraphics[width=\linewidth]{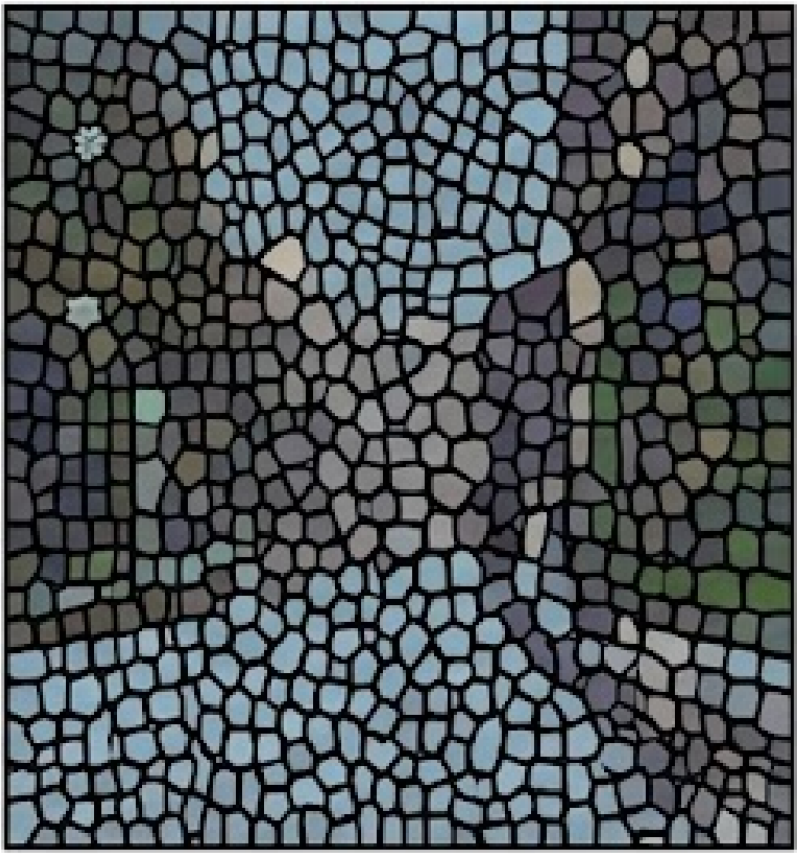}
        \\ \vspace{1mm}
        {\small (d) $K=32$}
    \end{minipage}
    
    \vspace{2mm}
    \caption{Visualization of the image partitioning with different values of $K$. As shown, $K=8$ leads to overly coarse and mixed semantic features, while $K=32$ results in fragmented regions. $K=16$ provides the most continuous and coherent semantic representation.}
    \label{fig:visual_k}
\end{figure}

\cref{tab:ablation_k} investigates the influence of the number of superpixels $K$ across different datasets, while the visual impact of region partitioning for different values of $K$ is illustrated in Fig.~\ref{fig:visual_k}. When $K$ is set to a smaller value ($K=8$), the image is partitioned into overly coarse regions, which limits the network's ability to accurately model fine-grained and spatially non-uniform degradation patterns. As $K$ increases to 16, the performance peaks across all benchmark datasets, achieving the highest average PSNR of 33.40 dB and SSIM of 0.9395. This indicates that an appropriate region partitioning granularity enables optimal structure-aware state propagation.
However, when $K$ is further increased to 32, the performance slightly degrades. Excessively large $K$ results in fragmented regions, diminishing the benefits of regional coherence and causing the scanning behavior to regress toward a pixel-wise traversal. Furthermore, as shown in \cref{tab:ablation_k}, a larger $K$ inevitably leads to higher overhead (e.g., FLOPs increases from 4.30G to 4.71G). Consequently, we adopt $K=16$ as the default setting for all experiments to achieve the best balance between restoration quality, regional structural integrity, and computational efficiency.

\begin{table}[t]
    \centering
    \caption{Comprehensive ablation studies of different scanning and gating strategies across four datasets. Default settings used in our final framework are highlighted in \textbf{bold}.}
    \label{tab:comprehensive_ablations}
    \scriptsize
    \setlength{\tabcolsep}{4pt} % 移除一列后稍微放开列间距，排版更美观
    \resizebox{\linewidth}{!}{
    \begin{tabular}{ll c c c c}
        \toprule
        & \textbf{Method} & \textbf{Outdoor} & \textbf{Snow100K-S} & \textbf{Snow100K-L} & \textbf{RainDrop} \\
        \midrule
        \multirow{4}{*}{\begin{tabular}[c]{@{}l@{}}\textbf{Scanning}\\ \textbf{Strategy}\end{tabular}} 
        & Raster Scan & 31.77 / 0.9130 & 36.53 / 0.9548 & 31.52 / 0.8823 & 31.82 / 0.9034 \\
        & Fixed Window Scan & 31.92 / 0.9215 & 36.78 / 0.9602 & 31.75 / 0.8904 & 31.94 / 0.9058 \\
        & Hilbert Scan & 32.03 / 0.9246 & 36.72 / 0.9557 & 31.79 / 0.8923 & 32.01 / 0.9067 \\
        & \textbf{Superpixel (Ours)} & \textbf{32.13 / 0.9288} & \textbf{36.93 / 0.9613} & \textbf{31.86 / 0.8946} & \textbf{32.16 / 0.9102} \\
        \midrule
        \multirow{3}{*}{\begin{tabular}[c]{@{}l@{}}\textbf{Gating}\\ \textbf{Strategy}\end{tabular}} 
        & w/o Gating & 32.13 / 0.9288 & 36.93 / 0.9613 & 31.86 / 0.8946 & 32.16 / 0.9102 \\
        & Global Gating & 32.91 / 0.9398 & 37.23 / 0.9608 & 32.19 / 0.9237 & 32.62 / 0.9396 \\
        & \textbf{Region Gating (Ours)} & \textbf{33.22 / 0.9401} & \textbf{37.87 / 0.9656} & \textbf{32.70 / 0.9297} & \textbf{32.82 / 0.9440} \\
        \bottomrule
    \end{tabular}
    }
\end{table}

\subsubsection{Effect of the Scanning Strategies.}
We investigate the impact of various scanning patterns to validate the proposed Superpixel-guided Selective Scan. Quantitative results across four benchmarks and visual partitioning are provided in \cref{tab:comprehensive_ablations} and Fig.~\ref{fig:visual_k}, respectively. Compared to conventional content-agnostic trajectories on the Outdoor dataset, such as Raster Scan (31.77 dB), Fixed Window Scan (31.92 dB), and Hilbert Scan (32.03 dB), our structure-aware scanning strategy achieves superior performance with 32.13 dB PSNR. While the superpixel clustering and permutation operations introduce a slight increase in computational cost (4.28G vs. 4.17G for Raster Scan), the mechanism successfully shifts the scanning paradigm from pixel-serial to a semantic-guided one. By partitioning the image into perceptually coherent regions and performing relations modeling within these semantic-related areas, our approach effectively prevents the model from learning non-discriminative features from semantic-conflict regions. This ensures that the state evolution adheres to semantic boundaries, leading to more effective restoration compared to rigid, predefined trajectories. This performance advantage consistently holds true across the other three weather datasets in \cref{tab:comprehensive_ablations}, where our superpixel scanning steadily delivers gains, outperforming the advanced Hilbert scanning by over 0.2 dB on the Snow100K-S benchmark.

\subsubsection{Effect of Region-Level Gating.} 
We evaluate the impact of the proposed Region-level Gating Mechanism (RGM) to verify its ability in performing intra-region semantic calibration. As shown in \cref{tab:comprehensive_ablations}, the "w/o Gating" baseline, which relies solely on the superpixel-guided scanning, yields 32.13 dB PSNR on the Outdoor dataset. While "Global Gating" reaches 32.91 dB, its image-wide statistics fail to capture the spatially non-uniform distribution of degradations. By transitioning to our proposed RGM (Region Gating), the performance is further boosted to the peak of 33.22 dB PSNR and 0.9401 SSIM. This improvement demonstrates that by adaptively modulating channel-wise feature responses based on regional semantic statistics rather than global ones, the model can more effectively emphasize restoration-critical information within each semantic unit while suppressing non-discriminative features from semantic-conflict regions. Notably, although RGM introduces a marginal increase in computational complexity (with an overhead of only 0.04G), this negligible cost leads to a significant performance gain of 1.09 dB PSNR over the baseline, confirming that regional calibration is essential for handling complex weather scenarios. As validated across the Snow100K-S, Snow100K-L, and RainDrop benchmarks in \cref{tab:comprehensive_ablations}, this regional strategy maintains a similar upward trend, where Region Gating improves upon Global Gating by over 0.5 dB on both the Snow100K-S and Snow100K-L datasets.

\subsection{Discussion and Limitations}

Although SSR achieves SOTA restoration performance with significantly reduced theoretical computational complexity, its practical inference speed warrants further discussion. As shown in \cref{tab:inference_time_flops_params}, while SSR is the fastest among Mamba-based models, it remains slightly slower than highly optimized Transformer counterparts. Unlike Transformers, which benefit from mature CUDA ecosystems and highly-fused kernels like FlashAttention, hardware-level acceleration for Mamba-based selective scan mechanisms remains in its infancy. Furthermore, our dynamic superpixel clustering and region-guided selective scanning require frequent memory permutations and irregular spatial indexing, leading to increased synchronization overheads on modern GPUs. Exploring customized, hardware-efficient CUDA kernels to mitigate these bottlenecks and fully unleash the linear-complexity potential of SSMs will be a key focus of our future work.

\begin{table}[t]
  \centering
  \caption{Inference latency (s) comparison among representative Mamba-based and Transformer-based architectures.}
  \label{tab:inference_time_flops_params}
  \setlength{\tabcolsep}{4pt}
  \footnotesize
  % 这里将原本的 C 替换为 c，确保 Venue 列和 Time 列都严格居中
  \begin{tabularx}{\textwidth}{c c c | c c c}
    \toprule
    \textbf{Transformer-based} & \textbf{Venue} & \textbf{Time} & \textbf{Mamba-based} & \textbf{Venue} & \textbf{Time} \\
    \midrule
    GridFormer  & IJCV'24 & 1.110 & MambaIR   & ECCV'24 & 2.599 \\
    Restormer   & CVPR'22 & 0.853 & MambaIRv2 & CVPR'25 & 1.918 \\
    SwinIR      & ICCVW'21& 0.823 & EVSSM     & CVPR'25 & 1.267 \\ 
    Histoformer & ECCV'24 & \textbf{0.716} & \textbf{SSR (Ours)} & \textbf{--} & \textbf{0.895} \\
    \bottomrule
  \end{tabularx}
\end{table}

Furthermore, the robustness of SSR inherently depends on the accuracy of the superpixel clustering. Since the segmentation process is content-dependent, extreme weather conditions (e.g., dense snow or torrential rain) can severely corrupt the underlying image structures. In such failure cases, the clustering algorithm may erroneously over-segment heavy degradations into independent semantic regions. This causes our semantic-guided trajectories to degrade into localized noise-fitting paths, thereby limiting the model's ability to leverage long-range clean contexts. Future research will explore degradation-robust clustering mechanisms and multi-scale semantic guidance to enhance the model's stability in severely corrupted scenarios.

\section{Conclusions}

In this paper, we propose \textbf{SSR}, a \textbf{S}emantic-center guilded \textbf{S}tate space model for image \textbf{R}estoration. 
Unlike conventional SSMs using rigid, content-agnostic scanning, SSR shifts the paradigm from pixel-serial to superpixel-based semantic-guided one.
By incorporating superpixel-guided structural cues via the proposed \textbf{$\text{S}^3$M}, our method ensures recursive state propagation is strictly confined within perceptually coherent regions, preventing the aggregation of semantically conflicting features across boundaries.
Furthermore, the \textbf{RGM} is developed to perform intra-region calibration by modulating degradation outliers within each semantic superpixel unit along the channel dimension.
Extensive experiments on six benchmarks demonstrate that SSR achieves SOTA performance with superior parameter efficiency. This work highlights semantic-aware SSMs as an efficient and interpretable alternative to complex attention-based designs for universal image restoration.

\section*{Acknowledgements}
This work was supported by the Shenzhen Science and Technology Program (No. JCYJ20240813114229039), PCL Major Key Project of PCL2025A17-2, the Natural Science Foundation of Tianjin, China (No. 24JCZXJC00040), the National Natural Science Foundation of China (No. 624B2072), the Doctoral Student Program of the Young S\&T Talents Cultivation Project, CAST, the Fundamental Research Funds for the Central Universities (No. 63263253), and the Supercomputing Center of Nankai University (NKSC).

\bibliographystyle{splncs04}
\bibliography{main}
\end{document}